\documentclass[11pt,a4paper]{article}

\usepackage{arxiv}
\usepackage[utf8]{inputenc}
\usepackage[T1]{fontenc}
\usepackage[english]{babel}
\usepackage{amssymb}
\usepackage{amsfonts}
\usepackage{amsmath}
\usepackage{dsfont} 
\usepackage{graphicx}
\usepackage{booktabs}
\usepackage{xcolor}
\usepackage{csquotes}
\usepackage{hyperref}
\usepackage[backend=biber, style=authoryear, natbib=true, datamodel=software]{biblatex}
\usepackage{software-biblatex}
\DeclareCiteCommand{\parencite}[\mkbibbrackets]
  {\usebibmacro{prenote}}%
  {\usebibmacro{citeindex}%
   \usebibmacro{cite}}
  {\multicitedelim}
  {\usebibmacro{postnote}}
\renewcommand*{\multicitedelim}{\addcomma\space}
\usepackage{array}
\DeclareSourcemap{
  \maps[datatype=bibtex]{
    \map{\step[fieldset=file, null]}
    \map{\step[fieldset=abstract, null]}
    \map{\step[fieldset=keywords, null]}
  }
}

\title{From Transcription to Semantic Corpus Analysis:\\
Unsupervised Learning of Sentence Representations for Ancient Languages}

\author{Théotime de la Selle\\
\small Institut des Sources Chrétiennes, HiSoMA, CNRS, Lyon, France}

\date{}

\begin{document}

\maketitle

\begin{abstract}
\noindent Automatic Text Recognition (ATR) now supplies digital humanities with large volumes of unstructured, heterogeneous, and often noisy text in ancient languages. Downstream semantic analyses---text reuse identification, alignment, and semantic search---rely on sentence embeddings, yet existing methods transfer poorly to ancient languages: generic multilingual encoders underperform, specialized language models yield anisotropic representation spaces, and labeled similarity data is scarce. We study two fully unsupervised strategies---TSDAE and unsupervised SimCSE (henceforth CSE)---that adapt a specialized token-level language model into a corpus-specific sentence encoder using only raw sentences. We validate them on biblical reuse in patristic literature in Latin and Ancient Greek: 2{,}936 sentences carrying expert-verified reuses in works of Augustine, Jerome, and Athanasius, matched against the Vulgate, the Septuagint, and the Nestle--Aland New Testament. We decompose reuse identification into binary detection and correspondence retrieval, evaluate both tasks separately, and benchmark the adapted encoders against multilingual, specialized, distilled, and supervised fine-tuned baselines, on clean text and on text perturbed by simulated HTR errors and scribal abbreviations. On these test sets, the adapted encoders lead all baselines on both tasks, with complementary profiles: TSDAE leads detection given a large in-domain corpus, while CSE leads retrieval, reaches its optimum with as few as 4--8k raw in-domain sentences---a few tens of seconds of training on a laptop GPU---and transfers across works and authors. Under simulated noise, retraining CSE on the perturbed text itself outperforms a clean-trained encoder. Measurements of alignment, uniformity, and anisotropy relate the geometric effect of each strategy to the measured gains. The full pipeline---segmentation, fine-tuning, cross-corpus semantic search---is available to non-specialists through the online tool Paraphrasis \citep{swh-dir-b0dd647}.
\end{abstract}


\section{Introduction}
\label{sec:intro}

The generalization of Automatic Text Recognition (ATR), and of Handwritten Text Recognition (HTR) in particular,\footnote{Throughout, ATR denotes the automatic transcription of both printed and handwritten sources; HTR is used only when handwriting is specifically meant, as in the simulated recognition errors of \autoref{sec:noise}.} has reshaped the data landscape of digital humanities \citep{muehlberger2019transkribus,kiessling2019escriptorium}. Scholars working on premodern sources now routinely produce machine-readable transcriptions at a scale that was unthinkable a decade ago. The scope of this transformation encompasses a wide range of applications, from the transcription of individual documents to the creation of extensive corpora\footnote{See for example CoMMA \citep{clerice:hal-05299220}}.
Following the pioneering phase of automatic handwriting recognition, a paradigm shift has emerged, and exploratory computational tools can now be designed and tested on large historical text databases. However, such transcribed handwritten corpora have a characteristic profile that differs from generic text databases and must be taken into account in analytical methods: \textit{lack of structuring}, \textit{HTR noise} from  recognition errors and \textit{heterogeneity} as linguistic, editorial, and completeness variability introduce systematic perturbations into the text (e.g. historical abbreviations or editorial artifacts). New philological tools have therefore to be developed and experimentally verified on historical datasets obtained by ATR.

Among all the new possible uses of texts, language representation models provide particularly promising text analysis methods. These neural networks based on \textit{Transformer} architecture \citep{vaswani2017attention} such as BERT models \citep{devlin2019bert} leverage deep learning processes to encode contextual meaning of words or sub-words into sets of numerical values, named \textit{semantic representations} or \textit{embeddings}. Using such embeddings, many recent research projects in the field of Natural Language Processing (NLP) have proposed algorithmic methods operating at the semantic level rather than purely syntactic or lexical levels \citep{qiu2020ptms}.\\

In the specific case of corpus analysis, semantic representations are particularly relevant at sequence-level i.e. clauses, sentences or text fragments as these so-called \textit{sentence embeddings} underpin a broad family of analyses that reduce to semantic operations between word sequences:
classification or alignment of texts \citep{liu_bertalign_2023};
paraphrase mining \citep{zhou_paraphrase_2025};
information retrieval (IR) \citep{karpukhin2020dpr};
statistical exploration through graphical projection and clustering \citep{grootendorst2022bertopic};
retrieval-augmented generation (RAG) \citep{lewis2020rag}; among others. Regarding cultural or historical corpus analysis, preliminary studies have attempted to adapt generic methods to philological fields for text alignment \citep{reboul_alignement_2024} or text reuse detection \citep{caffagni_benchmarking_2025}.
However, while this paradigm is well-established for modern, high-resource languages, three obstacles limit its application to premodern and ancient languages, especially after ATR where textual data from these languages exhibits high variability from available training datasets:
\begin{enumerate}
    \item \textbf{Generic multilingual encoders are inadequate:} Massively multilingual sentence encoders are trained predominantly on modern languages and transfer poorly to Latin or Ancient Greek, whose orthography, morphology, and domain vocabulary are underrepresented.
    \item \textbf{Specialized language models do not provide usable sequence representations out of the box:} Semantic representation of sequences can be obtained by token-level encoders trained for ancient languages \citep{bamman2020latinbert,riemenschneider2023exploring,cowenbreen2023logion}, but the resulting representations face underlying limitations (\autoref{sec:related}), that restrict the discriminative power of similarity-based analyses.
    \item \textbf{Labeled data is scarce:} Annotated semantic-similarity datasets, which drive supervised sentence-embedding training in high-resource settings \citep{reimers2019sbert,cer2017sts}, are essentially nonexistent for ancient languages. Expert-annotated intertextuality datasets now exist for a few works \citep{dexter2024database,caffagni_benchmarking_2025,mambelli2026biblical,schelb_loci_2026}, but extending supervision to a newly transcribed corpus requires new expert annotation.
\end{enumerate}

Therefore, the present work addresses an analytical step that follows transcription: enabling \emph{corpus semantic analysis} of such collections through learned semantic representations of textual sequences (clauses, sentences, or paragraphs) in ancient languages. Our response rests on three principles: (i)~adaptation to the specificities of a given corpus or document (lexical and typographic variation, recognition noise); (ii)~the adaptation of a token-level language model into a sequence encoder; and (iii)~fully unsupervised training, requiring only raw sentences and therefore no manual annotation. To validate and optimize the unsupervised learning strategies, we select a philologically central task: the identification of text reuse in Latin or ancient Greek corpus. More precisely, our experiments focus on identification of biblical reuses in patristic literature, as high quality and relatively large databases are available. Church Fathers quote, paraphrase, and allude to Scripture pervasively, and indexing these reuses is a long-standing scholarly endeavor \citep{mellerin2014biblindex}. Reuse in this case is frequently \emph{allusive}---paraphrastic and non-literal---which makes it a demanding testbed for semantic, rather than purely lexical, methods \citep{rel17010088, manjavacas2019allusive}.

The tested envelope should be stated at the outset. All experiments concern one genre of reuse---biblical quotation and allusion in quotation-dense patristic prose, matched against a closed canonical source corpus---in two languages, Latin and Ancient Greek, with reuse-bearing sentences making up 23\% to 50\% of the annotated works (\autoref{tab:corpora}).

This article makes five contributions.
\begin{enumerate}
    \item A training methodology based on two unsupervised strategies together with a study of their parameterization (training-corpus size and specialization, and noise-aware variants).
    \item A comparative analysis of sequence representations on two tasks that we argue should be evaluated separately: text reuse \emph{detection} (binary classification) and semantic correspondence \emph{retrieval} (Information Retrieval) against a source corpus.
    \item Experiments on several Latin and Ancient Greek datasets of biblical reuse in patristic literature, and on text perturbed by simulated HTR errors and abbreviations.
    \item An analysis of the effect of each training strategy on the geometry of the representation space (alignment, uniformity, anisotropy), complemented by UMAP atlases used as qualitative, exploratory views for philological work.
    \item An online ready-to-use tool, \textit{Paraphrasis} \citep{swh-dir-b0dd647}, enabling the automatic segmentation of raw texts, the fine-tuning of an existing model on provided sequences and the search of semantic correspondences linking two corpora through semantic similarity measures.    
\end{enumerate}

\section{Related Work}
\label{sec:related}

\paragraph{Sentence embeddings.} It has been widely observed that masked language models (e.g. BERT) \citep{devlin2019bert} perform poorly on sentence-level STS tasks owing to the anisotropy of the resulting embedding space \citep{ethayarajh2019contextual,kashyap_comprehensive_2024,li2020bertflow}. Two properties account for the superiority of dedicated sentence encoders \citep{wang2020alignment}: the \emph{alignment} of semantically related instances and the \emph{uniformity} of the representation distribution on the hypersphere. Building sequence-level semantic representations on top of Transformer encoders \citep{vaswani2017attention} was popularized by the Sentence-BERT framework \citep{reimers2019sbert}, which trains siamese networks on labeled natural-language-inference and similarity data. In the absence of labeled pairs of sequences, several unsupervised objectives have been proposed \citep{kashyap_comprehensive_2024}. Among all, two foundational sentence embedding learning strategies are proposed here and will be detailed in \autoref{sec:methods}: TSDAE \citep{wang2021tsdae} which trains a sequential denoising auto-encoder whose information bottleneck is the sentence embedding itself and SimCSE \citep{gao2021simcse} which applies contrastive learning \citep{oord2018cpc} with positive pairs generated by standard dropout, used as noise. Our work transfers these unsupervised objectives to ancient languages and studies their data-efficiency and noise-robustness in that regime, especially for philological tasks. Within the family of sentence-embedding methods, and to our knowledge, this is the first application of TSDAE to models of ancient languages and the first unsupervised SimCSE adaptation of Latin models. For Ancient Greek, \textsc{Krahn} et al. \citep{krahn_sentence_2023} trained SimCSE models as a baseline for their multilingual knowledge-distillation encoders, on Ancient Greek texts paired with English translations: SimCSE led on translation search but trailed distillation on semantic similarity and semantic retrieval. Our setting differs in that adaptation uses only monolingual raw sentences of the target corpus, with no translations. We show that contrastive adaptation substantially reduces the anisotropy of pre-trained encoders, whereas denoising auto-encoding accentuates it (\autoref{sec:disc_umap}).
\paragraph{Language models and sentence encoders for ancient languages.} Specialized masked language models exist for Latin---Latin BERT \citep{bamman2020latinbert} and LaBerta \citep{riemenschneider2023exploring}---and for Ancient Greek---GreBerta \citep{riemenschneider2023exploring} and Logion \citep{cowenbreen2023logion} but as explained above, their resulting sentence embeddings suffer from limitations. Sentence-level encoders for these languages remain comparatively rare. SPhilBerta \citep{riemenschneider_graecia_2023} is a trilingual (Ancient Greek, Latin, English) sentence encoder obtained by multilingual knowledge distillation \citep{reimers2020distillation}. The \textit{Loci Similes} benchmark and software \citep{schelb_loci_2026} provide sentence encoders for Latin intertextuality retrieval, obtained by fine-tuning bi-encoders that map each query and source segment independently into dense vectors with an Online Contrastive Loss on labeled parallels; their end-to-end pipeline couples a dense retrieval first stage with a cross-encoder classifier trained on labeled pairs. For biblical reuse in Latin patristic texts, \textsc{Mambelli} et al. \citep{mambelli2026biblical} fine-tune Latin BERT-based models with supervised contrastive learning and hard negative mining on Augustine's \textit{De Genesi ad litteram}. From another perspective, general-purpose multilingual sentence encoders \citep{feng2022labse,wang2024multilinguale5} can be used but cover ancient languages only marginally. We benchmark against the publicly released encoders among these and show that, on our test sets, lightweight corpus-specific unsupervised adaptation of a token-level model is more effective than off-the-shelf multilingual encoders, a distillation-based Latin/Greek encoder, and Latin sentence encoders fine-tuned for intertextuality on a supervised objective. In fact, consistently with reported language model behavior \citep{gururangan_dont_2020}, we demonstrate by varying the training corpus and experimented datasets that \textit{domain adaptation} can have a marked effect on the representations of sentences leading to accurate semantic similarity measures.     

\paragraph{Text reuse detection in the humanities.} Computational intertextuality has a long quantitative tradition in which systems are benchmarked against parallels attested by commentators or editors. Lexical approaches rely on shared word forms, lemmata, $n$-grams, and local alignment \citep{buechler2014textreuse}. The Tesserae project measured the recall of commentator-attested parallels between Lucan and Vergil \citep{coffee2012intertextuality,forstall2015modeling}; Tesserae matches have been aggregated to measure and map intergeneric allusion in Latin poetry \citep{burns2017measuring}; sequence alignment borrowed from bioinformatics has been applied to Latin intertextuality \citep{chaudhuri2017bioinformatics}; and \emph{passim} \citep{smith2014passim} remains a widely used system based on shared $n$-grams and alignment. For biblical material specifically, \textsc{Moritz} et al. \citep{moritz2016nonliteral} quantified the transformations undergone by Bible reuse in Greek and Latin authors and showed the limits of literal methods. Lexical approaches are effective for verbatim or near-verbatim borrowing but struggle with \emph{allusive} reuse, the paraphrastic, non-literal borrowing typical of patristic citation, for which semantic methods have been shown to be necessary \citep{rel17010088,manjavacas2019allusive}. Distributional semantics then entered the benchmarked setting: biblical allusion detection in Latin sermons was cast as retrieval with word embeddings \citep{manjavacas2019allusive}, and word-embedding methods were benchmarked against commentator-attested intertexts in Valerius Flaccus \citep{burns2021profiling}, whose 945 references were later published as a benchmark database \citep{dexter2024database}. Contextual models followed: suggestions computed with LatinBERT have been evaluated against attested allusions in Lucan \citep{gong2025augmented}, and sentence embeddings now support the detection of near-verbatim reuse, paraphrase, and allusion \citep{riemenschneider_graecia_2023,caffagni_benchmarking_2025, caffagni_generating_2025,taddei_detecting_2025,mambelli2026biblical,schelb_loci_2026}. Whatever the language---premodern or ancient---or literary field, text reuse detection studies based on sentence embeddings share a similar framework: each corpus is first segmented into sequences (clause, sentence, or paragraph). Each sequence is then encoded \emph{independently of the others} by a single model into a fixed-size vector, and all downstream operations---classification, retrieval, alignment---are carried out on these vectors via STS (often cosine similarity). Encoding sequences independently is what makes the approach scale: a corpus is embedded once, and any number of pairwise comparisons or nearest-neighbour queries can then be answered as vector operations. Within this sentence-embedding family, quantitative evaluations against expert-verified parallels remain few. For Latin, biblical reuse in Augustine's \textit{De Genesi ad litteram} has served to benchmark sentence encoders\footnote{The works of Caffagni et al. \citep{caffagni_benchmarking_2025, caffagni_generating_2025} use the same dataset of biblical reuses from Augustine's \textit{De Genesi ad litteram}.} \citep{caffagni_benchmarking_2025,caffagni_generating_2025} and supervised contrastive fine-tuning \citep{mambelli2026biblical}; the \textit{Loci Similes} benchmark \citep{schelb_loci_2026} links two Late Antique authors, Jerome and Lactantius, to ten classical Latin authors through 545 expert-verified parallels. For Ancient Greek, evaluations are qualitative \citep{riemenschneider_graecia_2023} or rely on synthetic parallels \citep{taddei_detecting_2025}. Our evaluation comprises 2{,}936 sentences carrying expert-verified biblical reuses in four Latin and Greek works; to our knowledge, it is the largest set used so far to evaluate sentence encoders on biblical reuse. Unlike the supervised approaches above, the adaptation studied here requires no labeled pairs, so that the labels serve only for evaluation. Our gold data derives from sustained philological projects indexing biblical quotations and allusions in early Christian literature \citep{mellerin2014biblindex}, encoded in XML-TEI \citep{tei2024guidelines}.

\section{Methods: unsupervised sentence representation learning}
\label{sec:methods}
As the goal of this work is to develop methodologies and experiments based on sentence representations dedicated to the semantic analysis of corpora obtained by ATR, we applied fully unsupervised sentence embedding learning on existing Latin and Ancient Greek models.   

\subsection{From token encoders to sequence encoders}
\label{sec:token2seq}

Our starting points are specialized masked language models, i.e.\ token-level encoders in Latin and Ancient Greek \citep{bamman2020latinbert,riemenschneider2023exploring,cowenbreen2023logion}. A sequence representation can be derived from such a model by pooling its token embeddings or by using the \textsc{[cls]} token, but, as noted in \autoref{sec:related}, the resulting space is anisotropic: representations are misaligned and non-uniformly distributed, which limits the usefulness of cosine similarity. The goal of the two strategies below is to reshape this space using only raw, unlabeled sentences drawn from the target domain. Crucially, both strategies are \emph{unsupervised}: they consume sentences, not annotations, and can therefore be applied to any corpus, including one that has just been transcribed.
\begin{figure}[htbp]
    \centering
    \includegraphics[width=1\linewidth]{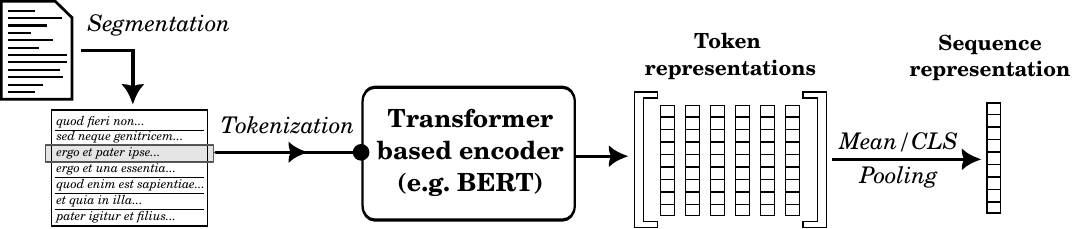}
    \caption{Architectures of the sequence-encoding models built on top of a pretrained token-level encoder.}
    \label{fig:pipeline}
\end{figure}

\subsection{TSDAE: denoising auto-encoding}
\label{sec:tsdae}

The first strategy is the Transformer-based Sequential Denoising Auto-Encoder \citep{wang2021tsdae}. It couples the pretrained encoder---in this case an existing Latin or Greek model---with a Transformer decoder in a denoising auto-encoder. Each sequence $s = (s_1, \dots, s_l)$, e.g. a sentence, is first corrupted---by deleting a fraction of its $l$ tokens---into $\tilde{s}$. Then, the encoder maps each corrupted sequence to a semantic representation $\mathbf{r}$, i.e. a sentence embedding, from which the decoder must reconstruct the \emph{original}, i.e. uncorrupted sequence $s$. Over a training corpus $\mathcal{D}$, training minimizes the cross-entropy of the autoregressive, token-by-token reconstruction, i.e. the negative log-likelihood of the original tokens, each predicted by a softmax over all the tokens of the vocabulary $V$:
\begin{equation}
\label{eq:tsdae}
\mathcal{L}_{\mathrm{TSDAE}}(\theta)
= -\,\mathbb{E}_{s \sim \mathcal{D}} \bigl[ \log P_\theta\bigl(s \mid \tilde{s}\bigr) \bigr]
= -\,\mathbb{E}_{s \sim \mathcal{D}} \Biggl[ \sum_{t=1}^{l} \log P_\theta\bigl(s_t \mid \mathbf{r},\, s_{<t}\bigr) \Biggr].
\end{equation}
Here $\theta$ denotes the set of all trainable parameters of the encoder--decoder network, which are progressively adjusted during training, and $\mathcal{L}_{\mathrm{TSDAE}}(\theta)$ is the \emph{loss}: a global penalty on reconstruction errors that training seeks to minimize by iteratively updating $\theta$. The notation $\mathbb{E}_{s \sim \mathcal{D}}[\cdot]$ (an \emph{expectation}) means that this penalty is averaged over sentences $s$ drawn from $\mathcal{D}$. The central term, $P_\theta(s \mid \tilde{s})$, reads as ``the probability that the network with weights $\theta$ assigns to the original sentence $s$ when shown only its corrupted version $\tilde{s}$''. This sentence-level probability decomposes into a sum over the $l$ tokens of $s$: at each decoding step $t$, $P_\theta(s_t \mid \mathbf{r},\, s_{<t})$ is the probability that the decoder predicts the correct token $s_t$, given the sequence vector $\mathbf{r}$ and the tokens $s_{<t}$ it has already reconstructed. In words: training adjusts the weights $\theta$ so that, on average over the corpus, the decoder assigns the highest possible probability to each token of the original sentence. After training, the decoder is discarded and only the encoder is retained.

\subsection{CSE: contrastive sentence embedding (unsupervised SimCSE)}
\label{sec:cse}

The second strategy follows unsupervised SimCSE \citep{gao2021simcse}; we call the resulting models \emph{CSE} models, and reserve the name SimCSE for the method of \textsc{Gao} et al. Each sequence is encoded twice under independent dropout masks, yielding a \emph{positive} pair of representations $\bigl(\mathbf{r}_i, \mathbf{r}_i^{+}\bigr)$, while the other sequences in the batch act as in-batch \emph{negatives}; the model is trained with the InfoNCE contrastive loss \citep{oord2018cpc} computed on cosine similarities. For a batch of $N$ sequences randomly selected within the training corpus, the loss for sequence $i$ is
\begin{equation}
\label{eq:infonce}
\ell_i = -\log \frac{\exp\bigl(\mathrm{sim}(\mathbf{r}_i, \mathbf{r}_i^{+})/\tau\bigr)}{\sum_{j=1}^{N} \exp\bigl(\mathrm{sim}(\mathbf{r}_i, \mathbf{r}_j^{+})/\tau\bigr)},
\end{equation}
where $\mathrm{sim}(\cdot,\cdot)$ denotes cosine similarity and $\tau$ a temperature hyper-parameter, i.e. a small constant that rescales the similarities and thereby controls how sharply the model is penalized for near-misses. The denominator ranges over the positive pair ($j = i$) and the $N-1$ in-batch negatives ($j \neq i$). Like Equation~\ref{eq:tsdae}, $\ell_i$ is a cross-entropy loss, minimized by adjusting the trainable parameters $\theta$ and averaged over all sentences of the batch.

\paragraph{Two cross-entropy objectives, two classification problems.} The two objectives therefore do not differ in form---both minimize a cross-entropy---but in the classification problem they pose and in where the sentence embedding enters it. TSDAE solves, at each decoding step, a $|V|$-way classification over the vocabulary; the embedding $\mathbf{r}$ enters only as the memory the decoder attends to, so that nothing in the loss constrains cosine similarities between embeddings directly. CSE solves an $N$-way classification over the sentences of the batch (instance discrimination), whose logits \emph{are} the cosine similarities $\mathrm{sim}(\mathbf{r}_i, \mathbf{r}_j^{+})/\tau$; the loss thus acts directly on the geometry that similarity search uses. Asymptotically, it optimizes the two properties identified by \textsc{Wang} and \textsc{Isola} \citep{wang2020alignment}: it pulls together representations of the same sequence under different dropout noise (alignment) and spreads representations over the hypersphere (uniformity). \autoref{sec:disc_umap} measures both properties on our encoders.

\subsection{Training details and configurations}
\label{sec:configs}

\paragraph{Hyper-parameters.} Unsupervised SimCSE models are trained following \textsc{Gao} et al. \citep{gao2021simcse} as presented above. Sentence embeddings are obtained by mean pooling over the token representations\footnote{Both mean and CLS pooling were tested during contrastive learning. Since experiments based on mean pooling always result in higher or equal values across all metrics, only mean pooling results are presented here.}. Starting from the Latin or Ancient Greek encoders, models are trained for a single epoch on various deduplicated Augustinian corpora (from 4k to $\sim$100k sentences) with a batch size of 32, a learning rate of 1e-5, a weight decay of 0.01, a linear warm-up over the first $6\%$ of training steps, a temperature $\tau = 0.05$, and a maximum sequence length of 256 tokens. TSDAE models follow the sequential denoising auto-encoder objective of \textsc{Wang} et al. \citep{wang2021tsdae} as presented above. Input sentences are corrupted by random token deletion (deletion ratio of 0.6, the recommended default), encoded into a fixed-size vector via CLS pooling. Training is run on various deduplicated patristic corpora (from 5k to $\sim$184k sentences) for three epochs with a batch size of 16, a learning rate of 2e-5 under a constant schedule with no warm-up, no weight decay, a maximum sequence length of 256 tokens, and mixed-precision (AMP) computation. After training, the decoder is discarded and only the encoder with CLS pooling is retained as the sentence embedding model. For both training strategies, other numbers of epochs were tested, resulting in either performance degradation or no notable change.

\paragraph{Variability and environment.} Training is stochastic (dropout masks, batch composition, token deletion), and each test set is a finite sample of sentences. All tables report a single training run. On the \textit{De Trinitate}, \autoref{app:variability} estimates both sources of variability: 95\% percentile bootstrap intervals over test sentences for every model \citep{efron1993bootstrap}, and the standard deviation over three training seeds for LaBerta-CSE\textsubscript{trin}; TSDAE, whose training on the full patristic corpus is far costlier, was trained once. Experiments were run on a single NVIDIA RTX PRO 4000 (Blackwell) laptop GPU with 16\,GB of memory; the software dependencies and their versions are listed in the \texttt{requirements.txt} file of the repository and of its Zenodo archive (\autoref{sec:conclusion}).

\paragraph{Naming convention.} Models are named after the base language model and the training strategy: LaBerta-TSDAE and LaBerta-CSE are obtained from LaBerta; Logion-CSE and GreBerta-CSE are the Greek counterparts. Subscripts denote the training corpus or variant: the training work (\textsubscript{trin} for the \textit{De Trinitate}, \textsubscript{cfaust} for the \textit{Contra Faustum}, \textsubscript{hier} for Jerome, \textsubscript{Aug} for the full Augustinian corpus); noise-aware training data (\textsubscript{HTR}, \textsubscript{ABV}; see~\autoref{sec:noise} and \autoref{app:noise}); or corpus size in sentences (e.g.\ LaBerta-CSE\textsubscript{6k} in \autoref{app:corpus_size}). Training on the raw sentences of the test work itself (e.g.\ LaBerta-CSE\textsubscript{trin} evaluated on the \textit{De Trinitate}) is transductive but uses no label: it corresponds to the intended use, adapting the encoder to the corpus under study, and the cross-work and cross-author variants of \autoref{sec:results_latin_gen} measure what is lost when this is not done. Unless a subscript indicates otherwise, TSDAE models are trained on a large patristic corpus ($\sim$184k deduplicated sentences) and CSE models on small, work- or author-specific corpora; the effect of corpus size is studied in \autoref{app:corpus_size}.

\section{Data: text reuse dataset, patristic and biblical texts}
\label{sec:data}

In order to validate and optimize unsupervised learning strategies presented in \autoref{sec:methods}, we propose several text reuse identification experiments, including non-literal and allusive references, which is philologically relevant to evaluate the quality of semantic representation. Originating mainly in the BiblIndex project\footnote{\url{https://www.biblindex.org/fr}}, our Latin or Ancient Greek texts as well as our text reuse dataset are composed of biblical quotations---verbatim or not---and allusions in patristic literature which have been identified and verified by experts.
\begin{table}[htbp]
    \centering
    \caption{Annotated patristic works used for training (sentence corpus) and experiments. Sources: CAG for Augustine; BibliText for Jerome (edition SC~637 \citep{SC637}) and Athanasius (edition SC~649 \citep{SC649}). \emph{Reuse sent.}: sentences carrying at least one expert-verified biblical reuse; \emph{Pos.}: proportion of such sentences, i.e.\ the positive rate of the detection task.}
    \label{tab:corpora}
   \begin{tabular}{llrrr}
    \toprule
    Author & Work & Sentences & Reuse sent. & Pos.\\
    \midrule
    \textsc{Augustine} & \textit{De Trinitate} & 4{,}410 & 1{,}189 & 27.0\%\\
    \textsc{Augustine} & \textit{Contra Faustum} & 4{,}694 & 1{,}069 & 22.8\%\\
    \textsc{Jerome} & \textit{Adversus Iovinianum} & 1{,}151 & 572 & 49.7\%\\
    \textsc{Athanasius} & \textit{De decretis Nicaenae synodi} & 393 & 106 & 27.0\%\\
    \midrule
    \multicolumn{2}{l}{Total} & 10{,}648 & 2{,}936 & 27.6\%\\
    \bottomrule
    \end{tabular}
\end{table}

The text reuse gold dataset is derived from digital editions, formatted in XML-TEI \citep{tei2024guidelines} and containing biblical reuses (quotations and allusions) encoded as XML segments, of various Church Fathers' works. \autoref{tab:corpora} lists the annotated works used for training---only sentences---and evaluation. Counts refer to sentences carrying at least one reuse (2{,}936 in total), not to reuses or verses, since a sentence may carry several reuses and a reuse may point to several verses. The four works span two authors writing in Latin (\textsc{Augustine of Hippo}, \textsc{Jerome of Stridon}) and one in Ancient Greek (\textsc{Athanasius of Alexandria}). Each document is first segmented into sentences based on punctuation and on word-count thresholds: a minimum of words to avoid unstable sentence embeddings \citep{wu2022esimcse} and a maximum of words to prevent signal dilution in extremely long Latin or Ancient Greek sentences. Second, each sentence containing one or more reuses, or part of one, is labeled as positive and associated with the corresponding biblical verses. Note that a single reuse may be associated with several biblical verses with different degrees of lexical or thematic proximity. The proportion of sentences carrying a reuse ranges from 22.8\% to 49.7\% (\autoref{tab:corpora}), so the detection task is moderately, not severely, imbalanced. The conditions of access to the texts and annotations are given in the data availability statement (\autoref{sec:conclusion}).

\begin{table}[htbp]
    \centering
    \caption{Source corpus: biblical verses, provided by the BiblIndex platform.}
    \label{tab:verses}
    \begin{tabular}{l>{\raggedright\arraybackslash}p{8.2cm}r}
    \toprule
    Version & Edition & Verses\\
    \midrule
    Latin & \textit{Stuttgart} or \textit{Weber--Gryson Vulgate} (VG) \citep{WeberGryson2007} & 37{,}227\\
    Ancient Greek & \textit{Rahlfs--Hanhart Septuagint} (LXX) \citep{RahlfsHanhart2006} and \textit{Novum Testamentum Graece} (NA28) \citep{NA28} & 37{,}457\\
    \bottomrule
    \end{tabular}
\end{table}

\autoref{tab:verses} presents the source corpus of biblical verses, provided by the BiblIndex's platform in tabular format (each sample corresponds to one verse). Pseudepigraphic texts such as the \textit{Odes} or the \textit{Psalms of Solomon} are also included.

\subsection{Simulated noise: HTR artifacts and abbreviations}
\label{sec:noise}

As historical corpora---especially from HTR processes---are heterogeneous, i.e. linguistic, editorial, and completeness variability introduce systematic perturbations into the text, we argue that the robustness of computational methods used in philological tools must be evaluated and discussed. Thus, to assess robustness without requiring a separately noisy gold corpus, we introduce an artificial perturbation relative to the standard Latin text with which baseline models are familiar, and compare adaptation on the perturbed text with adaptation on the same clean text. We simulate two types of perturbation: (i)~\emph{HTR-like artifacts}, which emulate the character confusions and spurious insertions/deletions produced by handwritten text recognition; and (ii)~\emph{abbreviations}, which emulate the contracted forms common in manuscripts and early prints. Both are synthetic, character- and token-level perturbations of the same clean edition, calibrated to a target character error rate or abbreviation rate (\autoref{app:noise}). They do not reproduce several properties of real recognition output: line-segmentation and reading-order errors, layout artifacts (running heads, marginalia, apparatus), and the systematic abbreviation-expansion policies of a given transcription model or guideline. The results of \autoref{sec:results_noise} are therefore properties of these simulated regimes, not tested properties of ATR pipelines; testing them on genuinely transcribed material is left for future work.

\section{Experiments: text reuse identification}
\label{sec:setup}

\subsection{Task decomposition}
\label{sec:tasks}
\begin{figure}[htbp]
    \centering
    \includegraphics[width=1\linewidth]{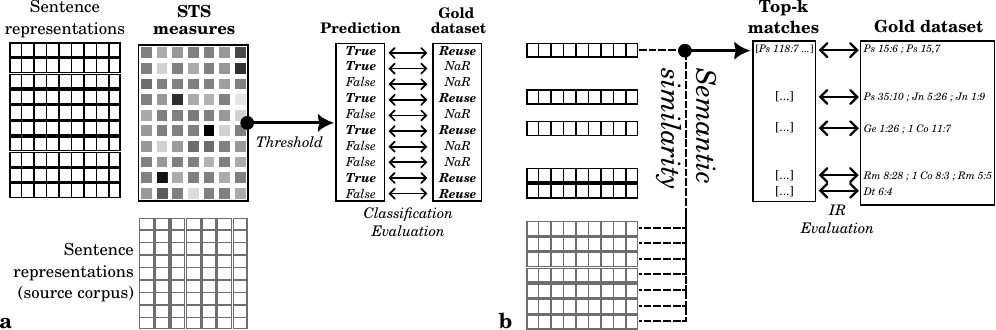}
    \caption{Schematic representation of text reuse identification decomposed in two algorithmic tasks: \textbf{a} binary \textit{classification} for text reuse detection and \textbf{b} \textit{information retrieval} (IR) for semantic search in the source corpus.}
    \label{fig:reuse_detection}
\end{figure}

Similarly to \textsc{Schelb} et al. \citep{schelb_loci_2026} we define text reuse identification, i.e. intertextuality detection, as the identification of directional dependencies between two texts: a query document (the chronologically later text) and a source document (the earlier text). The mapping between sentences of query document and source document is inherently one-to-many; a single query sentence may correspond to zero, one, or multiple distinct segments in the source text, exhibiting either semantic equivalence or stylistic imitation. Thus, rather than reporting a single figure of merit, we evaluate text reuse identification methods based on sentence representations on two algorithmic tasks (see \autoref{fig:reuse_detection}):

\begin{enumerate}
    \item \textbf{Reuse detection} (binary classification): \emph{does this sentence contain a reuse?} Each sentence is scored by its maximum similarity to any source verse, and the score distribution is evaluated against the gold labels (\autoref{fig:reuse_detection}\textbf{a}).
    \item \textbf{Correspondence retrieval} (semantic search / IR): \emph{which biblical verse does this sentence reuse?} For each sentence carrying a reuse, the source verses are ranked by similarity and the rank of the correct verse is examined (\autoref{fig:reuse_detection}\textbf{b}).
\end{enumerate}

Both component tasks rely on the same similarities, but they measure different things. The detection score $s_i = \max_{v} \mathrm{sim}(\mathbf{r}_i, \mathbf{r}_v)$ carries no information about \emph{which} verse attains the maximum: it can be high because of the gold verse, because of a verse parallel to it---the Bible repeats and quotes itself, as in synoptic parallels or Old Testament quotations in the New---or because of a generic proximity to biblical style and vocabulary. Detection therefore measures a necessary condition of identification, not identification itself. Retrieval measures localization conditional on the presence of a reuse. Neither component alone describes the quality an expert experiences when reviewing a model's output: a complete identification pipeline composes them, first flagging candidate sentences, then ranking candidate verses for each. Since the best detector and the best retriever need not be the same model (\autoref{sec:results}), the design and evaluation of such a two-stage pipeline are left for future work (\autoref{sec:conclusion}).

We nevertheless keep the two components separate, for three reasons. First, each corresponds to a distinct philological use. Detection alone serves triage: when indexing a newly transcribed corpus, it orders passages for expert review and estimates the density of scriptural language across a work. Retrieval alone serves the frequent case where a reuse is signaled but its source is not: introductory formulas (\textit{sicut scriptum est}, \textit{ait apostolus}), the typographic marking of quotations in printed editions, and the marginal \textit{diplai} by which late antique and early medieval scribes flagged quotations, often scriptural ones \citep{mcgurk1961citation}, all mark a quotation without identifying the verse; retrieval also serves the extension of a known reuse to further candidate sources. Second, the two components fail for different reasons and can be improved separately: a model may produce a well-separated score distribution (good detection) while ranking the exact source verse only moderately well (weaker retrieval), or vice versa (\autoref{sec:results}), and a composite score would hide which component limits a pipeline. Third, gold data rarely records intertextuality within the source corpus, which is dense and partly subjective in the Bible; this biases retrieval, where a parallel verse counts as an error, but not detection.

\subsection{Baseline models}
\label{sec:baselines}

We compare our adapted encoders against:
\begin{itemize}
    \item \textbf{Multilingual sentence encoders}: LaBSE \citep{feng2022labse} and multilingual-e5-base \citep{wang2024multilinguale5};
    \item \textbf{Specialized language models} used as sequence encoders by pooling: LaBerta and GreBerta \citep{riemenschneider2023exploring}, Latin BERT \citep{bamman2020latinbert}, and Logion \citep{cowenbreen2023logion};
    \item \textbf{A distilled classical-language sentence encoder}: SPhilBerta \citep{riemenschneider_graecia_2023};
    \item \textbf{Fine-tuned multilingual sentence encoders}: multilingual-e5-base-emb-lat-intertext-v1 and SPhilBerta-emb-lat-intertext-v1, released by the authors of the \textit{Loci Similes} benchmark \citep{schelb_loci_2026} and obtained by fine-tuning multilingual-e5-base and SPhilBerta with an Online Contrastive Loss on Latin parallel sentences\footnote{\textsc{Schelb} et al. report multilingual-e5-large as their most effective retriever; its fine-tuned version, multilingual-e5-large-emb-lat-intertext-v1, is also proposed, but we compare encoders with identical embedding dimensionality.}; 
    \item \textbf{Non-neural systems} (\autoref{app:results_other}): \emph{passim} \citep{smith2014passim}, with grid-search parameter optimization.
\end{itemize}
See \autoref{sec:related} for more details about these encoders. 

\subsection{Evaluation metrics}
\label{sec:metrics}

\paragraph{Detection.} Three metrics that require no threshold fixed in advance jointly probe the discriminative power of the distribution of similarity scores between sequence pairs.

\emph{Average Precision} (AP), sensitive to class imbalance,
\begin{equation}
\mathrm{AP} = \sum_{k} \bigl(R_k - R_{k-1}\bigr)\, P_k,
\end{equation}
where $P_k$ and $R_k$ are precision and recall at rank $k$. The chance level of AP equals the positive rate of the test set (22.8\% to 49.7\%, \autoref{tab:corpora}), so AP values are comparable only within a test set.

\emph{AUC-ROC} (Receiver Operating Characteristic--Area Under the Curve), invariant to class imbalance, the probability that a randomly drawn positive scores higher than a randomly drawn negative,
\begin{equation}
\text{AUC-ROC} = \Pr\bigl(s_i > s_j \mid y_i = 1,\, y_j = 0\bigr),
\end{equation}
where, for a sentence $i$, $y_i \in \{0,1\}$ is the gold label and $s_i \in [-1,1]$ the maximum similarity between $i$ and all verses.

\emph{$F1_{\max}$}, the maximum $F1$ score, i.e. harmonic mean of precision and recall, over thresholds,
\begin{equation}
\mathrm{F1}_{\max} = \max_{\tau}\, \frac{2\, P(\tau)\, R(\tau)}{P(\tau) + R(\tau)},
\end{equation}
where $P(\tau)$ and $R(\tau)$ are precision and recall at threshold $\tau$. Because the threshold is chosen on the test labels, $\mathrm{F1}_{\max}$ is an oracle (skyline) value: it upper-bounds the $F1$ attainable at a threshold fixed in advance. Like AP, it rises with the positive rate.

\paragraph{Retrieval.} \emph{Hits@$k$}, the proportion of queries for which at least one gold correspondence appears among the top-$k$ candidates,
\begin{equation}
\mathrm{Hits@}k = \frac{1}{|\mathcal{Q}|} \sum_{q \in \mathcal{Q}} \mathds{1}\bigl[\mathrm{rank}_q \leq k\bigr],
\end{equation}
where $\mathcal{Q}$ is the set of queries (the sentences carrying a reuse) and $\mathrm{rank}_q$ the rank of the first correct candidate for query $q$. Hits@$k$ is the metric closest to philological practice, where retrieved candidates are reviewed by experts and the question is whether the correct verse surfaces near the top of the list. It is, however, \emph{conditional}: it is computed only over sentences already known to carry a reuse, and therefore does not describe end-to-end quality, which depends on composing detection with retrieval. Result tables report $Hits@1$ and $Hits@10$.\footnote{We also computed the Mean Reciprocal Rank (MRR); as it ranked the models identically to $Hits@1$ on every test set, we omit it to avoid redundancy.}

\section{Results}
\label{sec:results}
For each sentence embedding model, we encode every biblical verse and every patristic segment into fixed-size, L2-normalized vectors. The biblical verse embeddings are indexed with a FAISS \citep{johnson_billion-scale_2017} \verb|IndexFlatIP| — an exact (non-approximate) inner-product index — which, since all vectors are L2-normalized, is equivalent to ranking verses by cosine similarity. 

\subsection{Benchmark of sequence representations}
\label{sec:results_benchmark}

\autoref{tab:trin_results} reports detection and retrieval performance on the \textit{De Trinitate}. Detection is evaluated on all sentences of the work; retrieval is evaluated on the $N = 1189$ sentences carrying a gold verse reference.

\begin{table}[htbp]
    \centering
    \caption{Text reuse detection ($AP$, $AUC$-$ROC$, $F1_{\max}$; computed from the maximum STS score per sentence) and correspondence retrieval against the \textit{Vulgate} ($Hits@1$, $Hits@10$; over the $N = 1189$ reuse sentences with gold verse references) on the \textit{De Trinitate} (\textsc{Augustine of Hippo}). Positive rate 27.0\% (1{,}189 of 4{,}410 sentences); AP and $F1_{\max}$ (an oracle value) are comparable only within a test set. Single training run; seed variability and confidence intervals in \autoref{app:variability}.}
    \label{tab:trin_results}
    \begin{tabular}{lrrr|rr}
    \toprule
    Model & $AP$ & $AUC$-$ROC$ & $F1_{\max}$ & $Hits@1$ & $Hits@10$\\
    \midrule
    LaBerta-TSDAE & \textbf{0.81}& \textbf{0.90} & \textbf{0.73} & 0.40 & 0.60\\
    LaBerta-CSE\textsubscript{trin} & 0.79 & 0.87 & 0.71 & \textbf{0.61} & \textbf{0.75}\\
    LaBerta & 0.72 & 0.82 & 0.62 & 0.42 & 0.56\\
    SPhilBerta-emb-lat-intertext-v1 & 0.66 & 0.79 & 0.58 & 0.43 & 0.59\\
    LatinBERT & 0.64 & 0.78 & 0.56 & 0.22 & 0.31\\
    SPhilBerta & 0.62 & 0.77 & 0.55 & 0.39 & 0.56\\
    multilingual-e5-base-emb-lat-intertext-v1 & 0.65 & 0.76 & 0.56 & 0.48 & 0.64\\
    multilingual-e5-base & 0.60 & 0.73 & 0.52 & 0.37 & 0.50\\
    LaBSE & 0.52 & 0.70 & 0.50 & 0.30 & 0.41\\
    \bottomrule
    \end{tabular}
\end{table}

\paragraph{Detection task.} On detection (\autoref{tab:trin_results}, left-hand columns), the two adapted encoders largely lead on every metric: LaBerta-TSDAE attains the best scores throughout, with LaBerta-CSE\textsubscript{trin} close behind: $0.02$ to $0.03$ lower on every detection metric, a gap of the same order as the standard deviations over training seeds (\autoref{app:variability}). Both markedly outperform the pooled representations of the base language model LaBerta ($+0.11$ and $+0.09$ in $F1_{\max}$, $+0.09$ and $+0.07$ in $AP$, for LaBerta-TSDAE and LaBerta-CSE\textsubscript{trin} respectively), and the gap to the generic multilingual encoders (LaBSE and multilingual-e5-base) remains wide. The oracle $F1_{\max}$ separates four groups: the two adapted encoders ($0.71$--$0.73$); the unadapted LaBerta ($0.62$); LatinBERT, SPhilBerta, and the two encoders fine-tuned for intertextuality ($0.55$--$0.58$); and the generic multilingual encoders ($0.50$--$0.52$). The 95\% bootstrap intervals (\autoref{app:variability}) separate both adapted encoders from LaBerta on every detection metric, and LaBerta-TSDAE from LaBerta-CSE\textsubscript{trin} on $AUC$-$ROC$ only.

\paragraph{IR task.} Retrieval (right-hand columns) partly inverts this picture and separates the models far more sharply. LaBerta-CSE\textsubscript{trin} dominates both rank-based metrics ($Hits@1$ $0.61$), with a large margin over the second-best system ($Hits@1$ $0.48$); in practical terms, its $Hits@10$ of $0.75$ means that for roughly three quarters of the reuse sentences, the gold verse appears among the ten candidates an expert would review. LaBerta-TSDAE, the best detector, falls to fifth place in $Hits@1$ ($0.40$)---just below the unadapted LaBerta ($0.42$)---which is the starkest evidence in our results that detection and retrieval measure different capabilities (\autoref{sec:disc_tasks}). LaBerta-TSDAE, however, gains more from rank 1 to rank 10 ($+0.20$) than LaBerta ($+0.14$): it falls $0.02$ below its base in $Hits@1$ but exceeds it by $0.04$ in $Hits@10$. This larger gain from rank 1 to rank 10 is also found on the \textit{Contra Faustum} and under both noise regimes (Tables~\ref{tab:cfaust_results}, \ref{tab:htr_results}, \ref{tab:abv_results}). For review lists of ten candidates, both adapted encoders therefore improve on their base, CSE by a wide margin. Four further observations stand out regarding the IR task.
\begin{itemize}
    \item \textbf{The fine-tuned sentence encoders of \citep{schelb_loci_2026}} (suffix \textit{-emb-lat-intertext-v1}) improve consistently over their base encoders on both tasks. For multilingual-e5-base, the gain is larger in retrieval than in detection: $+0.11$ in $Hits@1$ ($0.37 \rightarrow 0.48$), against $+0.05$ in $AP$ ($0.60 \rightarrow 0.65$); supervised learning with an Online Contrastive Loss on Latin parallel sentences thus chiefly reshapes the \emph{ranking} behaviour of its space. For SPhilBerta, the gains are smaller and similar on both tasks ($Hits@1$ $0.39 \rightarrow 0.43$, $AP$ $0.62 \rightarrow 0.66$). Since these encoders were fine-tuned on parallels between Late Antique and classical authors rather than on biblical reuse, their comparison with our encoders mixes the effect of supervision with that of domain.
    \item \textbf{LaBSE and multilingual-e5-base} confirm that generic multilingual sequence encoders are inadequate for Latin: both fall below the pooled representations of the monolingual language model on nearly every metric of both tasks. 
    \item \textbf{SPhilBerta}, the trilingual sentence encoder obtained by cross-lingual knowledge distillation, scores below the unadapted LaBerta in detection as in retrieval. Our design does not disentangle the causes, since SPhilBerta differs from LaBerta in three respects: its base model (the trilingual PhilBERTa), its objective (a mean-squared distillation loss that maps a Latin sentence onto an English teacher's embedding of its translation), and its training data (translation pairs). Because these pairs include biblical text \citep{riemenschneider_graecia_2023}, a lack of exposure to biblical Latin is unlikely to be the sole explanation. The objective is a plausible factor: it rewards translation equivalence across languages, a stricter and different relation than monolingual paraphrase or allusion.\item \textbf{LatinBERT} is competitive in detection ($AP$ $0.64$, above SPhilBerta) but collapses in retrieval, ranking last on both rank-based metrics ($Hits@1$ $0.22$, $Hits@10$ $0.31$)---a collapse that recurs in every subsequent Latin experiment (not displayed in subsequent tables for brevity).
\end{itemize}
For brevity, we have excluded three less competitive models (LaBSE, multilingual-e5-base and LatinBERT) from the following tables related to Latin datasets as their behaviors stay similar in all Latin experiments. 

\subsection{Generalization to other Latin works and authors}
\label{sec:results_latin_gen}

Tables~\ref{tab:cfaust_results} and~\ref{tab:hier_results} extend the evaluation to a second Augustinian work, the \textit{Contra Faustum}, and to a different author, respectively. The \textit{Contra Faustum} experiment now includes a third contrastive variant, LaBerta-CSE\textsubscript{hier}, trained on Jerome's \textit{Adversus Iovinianum}, which allows transfer to be assessed across works \emph{and} across authors on the same test set.
\begin{table}[htbp]
    \centering
    \caption{Text reuse detection ($AP$, $AUC$-$ROC$, $F1_{\max}$; computed from the maximum STS score per sentence) and correspondence retrieval against the \textit{Vulgate} ($Hits@1$, $Hits@10$; over the $N = 1069$ reuse sentences with gold verse references) on the \textit{Contra Faustum} (\textsc{Augustine of Hippo}). Positive rate 22.8\% (1{,}069 of 4{,}694 sentences); AP and $F1_{\max}$ (an oracle value) are comparable only within a test set. Single training run; variability is estimated on the \textit{De Trinitate} (\autoref{app:variability}).}
    \label{tab:cfaust_results}
    \begin{tabular}{lrrr|rr}
    \toprule
    Model & $AP$ & $AUC$-$ROC$ & $F1_{\max}$ & $Hits@1$ & $Hits@10$\\
    \midrule
    LaBerta-TSDAE & \textbf{0.76} & \textbf{0.87} & 0.67 & 0.41 & 0.58\\
    LaBerta-CSE\textsubscript{cfaust} & \textbf{0.76} & 0.86 & 0.68 & \textbf{0.60} & \textbf{0.74}\\
    LaBerta-CSE\textsubscript{trin} & \textbf{0.76} & 0.85 & \textbf{0.69} & \textbf{0.60} & \textbf{0.74}\\
    LaBerta-CSE\textsubscript{hier} & 0.74 & 0.84 & 0.67 & 0.59 & \textbf{0.74}\\
    multilingual-e5-base-emb-lat-intertext-v1 & 0.66 & 0.79 & 0.58 & 0.51 & 0.66\\
    LaBerta & 0.65 & 0.79 & 0.56 & 0.45 & 0.57\\
    SPhilBerta-emb-lat-intertext-v1 & 0.63 & 0.79 & 0.54 & 0.46 & 0.60\\
    SPhilBerta & 0.57 & 0.76 & 0.50 & 0.39 & 0.56\\
    \bottomrule
    \end{tabular}
\end{table}

\paragraph{Detection task.} The \textit{De Trinitate} pattern is confirmed and sharpened. LaBerta-TSDAE shares the best $AP$ ($0.76$) and keeps the best $AUC$-$ROC$ ($0.87$), although the best $F1_{\max}$ now goes to a contrastive model ($0.69$, LaBerta-CSE\textsubscript{trin}); the three contrastive variants and TSDAE all clearly outperform the fine-tuned multilingual sentence encoders and the unadapted LaBerta, which obtain similar values. 

\paragraph{IR task.} The two Augustinian CSE variants obtain equal scores and the Hieronymian CSE variant is just below, while LaBerta-TSDAE is positioned around the unadapted LaBerta ($-0.04$ in $Hits@1$, $+0.01$ in $Hits@10$) and below the fine-tuned multilingual sentence encoders---making the detection/retrieval dissociation starker still than on the \textit{De Trinitate}. The fine-tuned multilingual-e5 variant is again a strong retriever ($Hits@1$ $0.51$, fourth overall) despite mid-table detection. 

Considering both tasks, the transfer result is the most consequential: relative to in-work training, the cross-work, same-author model LaBerta-CSE\textsubscript{trin} obtains almost equal results as LaBerta-CSE\textsubscript{cfaust}, and the cross-author model LaBerta-CSE\textsubscript{hier}, trained on Jerome, loses about one point---a graceful degradation with domain distance (work $\gtrsim$ author) that leaves all three far above every non-contrastive alternative.

\begin{table}[htbp]
    \centering
    \caption{Text reuse detection ($AP$, $AUC$-$ROC$, $F1_{\max}$; computed from the maximum STS score per sentence) and correspondence retrieval against the \textit{Vulgate} ($Hits@1$, $Hits@10$; over the $N = 572$ reuse sentences with gold verse references) on the \textit{Adversus Iovinianum} (\textsc{Jerome of Stridon}). Positive rate 49.7\% (572 of 1{,}151 sentences); AP and $F1_{\max}$ (an oracle value) are comparable only within a test set. Single training run; variability is estimated on the \textit{De Trinitate} (\autoref{app:variability}).}
    \label{tab:hier_results}
    \begin{tabular}{lrrr|rr}
    \toprule
    Model & $AP$ & $AUC$-$ROC$ & $F1_{\max}$ & $Hits@1$ & $Hits@10$\\
    \midrule
    LaBerta-TSDAE & \textbf{0.92} & \textbf{0.91} & \textbf{0.84} & 0.52 & 0.66\\
    LaBerta-CSE\textsubscript{hier} & 0.90 & 0.88 & 0.79 & \textbf{0.67} & \textbf{0.81}\\
    LaBerta-CSE\textsubscript{trin} & 0.89 & 0.87 & 0.79 & 0.64 & 0.80\\
    LaBerta & 0.88 & 0.86 & 0.79 & 0.51 & 0.65\\
    SPhilBerta-emb-lat-intertext-v1 & 0.85 & 0.84 & 0.76 & 0.53 & 0.70\\
    multilingual-e5-base-emb-lat-intertext-v1 & 0.85 & 0.82 & 0.74 & 0.57 & 0.73\\
    SPhilBerta & 0.82 & 0.80 & 0.73 & 0.46 & 0.64\\
    \bottomrule
    \end{tabular}
\end{table}

\paragraph{Detection task.} On the \textit{Adversus Iovinianum} (\autoref{tab:hier_results}), detection scores are uniformly higher than on the Augustinian works and the unadapted LaBerta becomes nearly indistinguishable from the contrastive variants on the three metrics. One property of the test set should temper cross-work comparison here: reuse density is much higher (572 of 1{,}151 sentences, i.e.\ $49.7\%$ positives, against $27.0\%$ on the \textit{De Trinitate}), which mechanically raises the baseline of the prevalence-sensitive metrics $AP$ and $F1_{\max}$; $AUC$-$ROC$, being prevalence-invariant (\autoref{sec:metrics}), remains the appropriate metric for comparison across works. On that metric the ordering and absolute values are essentially unchanged: LaBerta-TSDAE leads ($0.91$), followed by the contrastive variants, then the fine-tuned multilingual sentence encoders---again above their respective bases but below LaBerta. 

\paragraph{IR task.} The CSE variants again lead both rank-based metrics: LaBerta-CSE\textsubscript{hier} reaches $Hits@1$ $0.67$, and its $Hits@10$ of $0.81$ means that the gold verse appears among the top ten candidates for over four fifths of the queries. The fine-tuned multilingual sentence encoders once more outrank their bases---and even LaBerta-TSDAE, which here edges marginally above the unadapted LaBerta on $Hits@1$ and $Hits@10$. LaBerta-CSE\textsubscript{trin}, trained on Augustine, transfers to Jerome with a loss of three points ($Hits@1$ $0.64$ vs.\ $0.67$ for in-work training).

Taken together, the two test sets show that the contrastive adaptation captures regularities of patristic Latin that extend beyond the specific training document---and beyond its author---while a small amount of in-work adaptation still yields a measurable additional gain.

\subsection{Extension to Ancient Greek}
\label{sec:results_greek}

\autoref{tab:greek_results} transposes the methodology to Ancient Greek on the \textit{De decretis Nicaenae synodi}, starting from two Greek language models, Logion \citep{cowenbreen2023logion} and GreBerta \citep{riemenschneider2023exploring}. We omit Ancient-Greek-BERT \citep{ancient-greek-bert} from the comparison. Logion \citep{cowenbreen2023logion} is a BERT model for premodern Greek trained on roughly 70M words, about ten times the 6.4M-word corpus assembled for Ancient-Greek-BERT. The model described in \citep{cowenbreen2023logion} used the tokenizer of Ancient-Greek-BERT, itself built for Modern Greek subwords, and its authors found that training it from random initialization or from Ancient-Greek-BERT led to the same validation loss. The checkpoint we use, \texttt{princeton-logion/logion-bert-base}\footnote{\url{https://huggingface.co/princeton-logion/logion-bert-base}}, belongs to a later release whose WordPiece tokenizer, with a vocabulary of about 50k tokens, was trained on the same corpus of over 70M words: it therefore does not share the vocabulary of Ancient-Greek-BERT, nor, consequently, its embedding layer. Since it is trained on a much larger premodern Greek corpus, with a vocabulary built for it, we consider that it supersedes Ancient-Greek-BERT for this comparison. The representative contrastive model Logion-CSE\textsubscript{6k} is trained on $6k$ sentences from \textsc{Athanasius}'s texts. Retrieval is evaluated over the $N = 106$ reuse sentences with gold verse references. No TSDAE version is proposed, for lack of a sufficiently large coherent corpus of Greek patristic sentences.\footnote{As reported in \autoref{app:corpus_size}, the performance of the Latin TSDAE model decreases sharply as its training corpus shrinks: on the \textit{De Trinitate}, its $AUC$-$ROC$ falls from $0.90$ with $\sim$184k sentences to $0.86$ with 40k and $0.62$ with 5k, below the unadapted LaBerta ($0.82$).}

\begin{table}[htbp]
    \centering
    \caption{Text reuse detection ($AP$, $AUC$-$ROC$, $F1_{\max}$; computed from the maximum STS score per sentence) and correspondence retrieval against the Greek biblical reference corpus (\textit{Septuagint} and NA28; $Hits@1$, $Hits@10$; over the $N = 106$ reuse sentences with gold verse references) on the \textit{De decretis Nicaenae synodi} (\textsc{Athanasius of Alexandria}). Positive rate 27.0\% (106 of 393 sentences); AP and $F1_{\max}$ (an oracle value) are comparable only within a test set. Single training run; variability is estimated on the \textit{De Trinitate} (\autoref{app:variability}).}
    \label{tab:greek_results}
    \begin{tabular}{lrrr|rr}
    \toprule
    Model & $AP$ & $AUC$-$ROC$ & $F1_{\max}$ & $Hits@1$ & $Hits@10$\\
    \midrule
    Logion-CSE\textsubscript{6k}& \textbf{0.74} & \textbf{0.84} & \textbf{0.68} & 0.40 & \textbf{0.58}\\
    Logion & 0.69 & 0.81 & 0.60 & 0.28 & 0.38\\
    GreBerta-CSE\textsubscript{6k}& 0.68 & 0.80 & 0.64 & \textbf{0.41} & \textbf{0.58}\\
    GreBerta & 0.51 & 0.71 & 0.52 & 0.26 & 0.36\\
    SPhilBerta & 0.48 & 0.71 & 0.51 & 0.22 & 0.32\\
    multilingual-e5-base & 0.50 & 0.65 & 0.47 & 0.20 & 0.31\\
    LaBSE & 0.40 & 0.57 & 0.45 & 0.12 & 0.22\\
    \bottomrule
    \end{tabular}
\end{table}

\paragraph{Detection task.} Each CSE adaptation improves on its base on every metric. GreBerta-CSE\textsubscript{6k} shows the larger improvement ($+0.17$ in $AP$, against $+0.05$ for Logion-CSE\textsubscript{6k}), but the Logion-based contrastive model leads every detection metric, because the base GreBerta detects far worse than Logion. The multilingual encoders trail the field, as in Latin. 

\paragraph{IR task.} The best systems are the contrastive variants, GreBerta-CSE\textsubscript{6k} and Logion-CSE\textsubscript{6k}; with $N = 106$ queries, their one-point difference amounts to a single query and is within sampling noise. Both far exceed their bases and the multilingual models. Absolute retrieval scores are lower than in Latin, but the gains of the CSE encoders over their base models are of similar magnitude ($+0.12$ and $+0.15$ in $Hits@1$, against $+0.15$ to $+0.19$ in Latin). The lower absolute scores thus mirror the lower scores of the Greek base models. A smaller pre-training corpus does not explain them by itself, since the Greek corpus of GreBerta (185.1M tokens) is larger than the Latin corpus of LaBerta (167.5M tokens) \citep{riemenschneider2023exploring}; with a single Greek work, our design cannot separate the effect of the models from that of the test work and its source corpus.

The Greek results thus replicate the Latin pattern for CSE, and the near-identical performance of the Logion-CSE variants across training sizes (\autoref{app:corpus_size}) mirrors the Latin saturation finding. With a single Greek work and 106 reuse sentences, this is a replication on a small test set; it suggests that the gains are not specific to Latin.

\subsection{Robustness to simulated noise}
\label{sec:results_noise}

Tables~\ref{tab:htr_results} and~\ref{tab:abv_results} report performance on \textit{De Trinitate} sentences perturbed by the two simulated regimes, HTR-like artifacts and abbreviations (\autoref{sec:noise}); each regime is evaluated on detection and on retrieval over the $N = 1189$ reuse sentences with gold verse references.

\begin{table}[htbp]
    \centering
    \caption{Text reuse detection ($AP$, $AUC$-$ROC$, $F1_{\max}$; computed from the maximum STS score per sentence) and correspondence retrieval against the \textit{Vulgate} ($Hits@1$, $Hits@10$; over the $N = 1189$ reuse sentences with gold verse references) on the \textit{De Trinitate} with simulated HTR artifacts. Positive rate 27.0\%; AP and $F1_{\max}$ (an oracle value) are comparable only within a test set. Single training run; variability is estimated on the \textit{De Trinitate} (\autoref{app:variability}).}
    \label{tab:htr_results}
    \begin{tabular}{lrrr|rr}
    \toprule
    Model & $AP$ & $AUC$-$ROC$ & $F1_{\max}$ & $Hits@1$ & $Hits@10$\\
    \midrule
    LaBerta-TSDAE & \textbf{0.54}& \textbf{0.70} & \textbf{0.54} & 0.12 & 0.21\\
    LaBerta-CSE\textsubscript{HTR} & \textbf{0.54} & 0.67 & 0.49 & \textbf{0.24} & \textbf{0.36}\\
    LaBerta-CSE\textsubscript{trin} & 0.49 & 0.63 & 0.45 & 0.20 & 0.29\\
    SPhilBerta-emb-lat-intertext-v1 & 0.43 & 0.63 & 0.46 & 0.15 & 0.23\\
    LaBerta & 0.45 & 0.62 & 0.45 & 0.11 & 0.16\\
    SPhilBerta & 0.39 & 0.61 & 0.45 & 0.11 & 0.20\\
    multilingual-e5-base-emb-lat-intertext-v1 & 0.37 & 0.56 & 0.43 & 0.17 & 0.25\\
    \bottomrule
    \end{tabular}
\end{table}

\begin{table}[htbp]
    \centering
    \caption{Text reuse detection ($AP$, $AUC$-$ROC$, $F1_{\max}$; computed from the maximum STS score per sentence) and correspondence retrieval against the \textit{Vulgate} ($Hits@1$, $Hits@10$; over the $N = 1189$ reuse sentences with gold verse references) on the \textit{De Trinitate} with simulated abbreviations. Positive rate 27.0\%; AP and $F1_{\max}$ (an oracle value) are comparable only within a test set. Single training run; variability is estimated on the \textit{De Trinitate} (\autoref{app:variability}).}
    \label{tab:abv_results}
    \begin{tabular}{lrrr|rr}
    \toprule
    Model & $AP$ & $AUC$-$ROC$ & $F1_{\max}$ & $Hits@1$ & $Hits@10$\\
    \midrule
    LaBerta-TSDAE & 0.76 & \textbf{0.88} & \textbf{0.70} & 0.27 & 0.42\\
    LaBerta-CSE\textsubscript{ABV} & \textbf{0.78} & 0.86 & \textbf{0.70} & \textbf{0.53} & \textbf{0.70}\\
    LaBerta-CSE\textsubscript{trin} & 0.75 & 0.83 & 0.66 & 0.51 & 0.68\\
    LaBerta & 0.66 & 0.79 & 0.58 & 0.31 & 0.44\\
    SPhilBerta-emb-lat-intertext-v1 & 0.62 & 0.77& 0.56 & 0.36 & 0.52\\
    SPhilBerta & 0.58 & 0.75 & 0.54 & 0.31 & 0.46\\
    multilingual-e5-base-emb-lat-intertext-v1 & 0.54 & 0.70 & 0.47 & 0.40 & 0.56\\
    \bottomrule
    \end{tabular}
\end{table}

Simulated noise degrades every system, and retrieval suffers far more than detection: under HTR-like artifacts the best $Hits@1$ falls to $0.24$, against $0.61$ on clean text. Abbreviations are markedly better tolerated (best $Hits@1$ $0.53$; best $AUC$-$ROC$ $0.88$, against $0.70$ under HTR artifacts), which is intuitive, since abbreviation mostly shortens tokens whereas HTR errors scramble characters more disruptively. Three observations stand out.
\begin{itemize}
    \item \textbf{Noise-aware training pays on both tasks.} A CSE model retrained directly on the noised sentences (LaBerta-CSE\textsubscript{HTR}, LaBerta-CSE\textsubscript{ABV}) beats the clean-trained LaBerta-CSE\textsubscript{trin} on \emph{every} metric under both regimes---in retrieval, $Hits@1$ rises from $0.20$ to $0.24$ under HTR noise and from $0.51$ to $0.53$ under abbreviations, and under abbreviations LaBerta-CSE\textsubscript{ABV} is by far the strongest retriever, with the largest IR improvement over the base model among all CSE adaptations in this article ($+0.22$ in $Hits@1$ and $+0.26$ in $Hits@10$); in detection, the noise-aware models take the best $AP$ under abbreviations ($0.78$) and tie with LaBerta-TSDAE under HTR noise ($0.54$).
    \item \textbf{The detection/retrieval dissociation persists under noise.} LaBerta-TSDAE keeps the best $AUC$-$ROC$ under both regimes yet sinks to the bottom tier in retrieval---below the unadapted LaBerta under abbreviations ($Hits@1$ $0.27$ vs.\ $0.31$), and only marginally above it under HTR noise.
    \item \textbf{Remediation strategy.} Given how light CSE training is (\autoref{app:corpus_size}), these results point to a concrete strategy: rather than only cleaning the transcription, one can adapt the encoder to the actual, imperfect state of the transcribed corpus. This holds for the simulated regimes tested here; whether it holds for real recognition output, with its segmentation, layout, and expansion errors, remains to be tested (\autoref{sec:noise}).
\end{itemize}

\section{Discussion}
\label{sec:discussion}

\subsection{Separating detection from retrieval}
\label{sec:disc_tasks}

Our results support keeping detection and retrieval methodologically distinct, and evaluating models primarily on retrieval, which proves the more discriminating of the two. Detection metrics rank LaBerta-TSDAE first on the \textit{De Trinitate} (\autoref{tab:trin_results}), yet on retrieval it falls to fifth place ($Hits@1$ rank), just below the unadapted LaBerta, while LaBerta-CSE\textsubscript{trin} retrieves the correct verse at rank one over twenty points more often ($Hits@1$ $0.61$ vs.\ $0.40$). A single aggregate detection score would therefore hide the model best suited to the end use, where candidate verses are reviewed by experts. The retrieval metrics also spread the field far more widely than the detection metrics---$Hits@1$ spans $0.22$--$0.61$ where $AUC$-$ROC$ spans $0.70$--$0.90$---and they reorder it: the intertext-adapted multilingual variant overtakes the unadapted Latin language models and LaBerta-TSDAE in ranking quality while remaining mid-field in detection. The dissociation recurs on every Latin test set (Tables~\ref{tab:cfaust_results}, \ref{tab:hier_results}, \ref{tab:htr_results}, \ref{tab:abv_results}): the $AUC$-$ROC$ leader, LaBerta-TSDAE, is never the retrieval leader, and on the \textit{Contra Faustum} and under simulated noise it sinks to the bottom tier in retrieval, around or below the unadapted LaBerta. This is a methodological point of general relevance to computational text-reuse studies, which often report a single figure of merit. Since the best detector and the best retriever are different models, a complete identification pipeline would not reduce to a single encoder: coupling a detector with a retriever, and evaluating the pair end to end, is left for future work (\autoref{sec:conclusion}).

\subsection{TSDAE versus CSE: data regimes, generalization and domain adaptation }
\label{sec:disc_methods}

The two strategies occupy complementary niches. TSDAE, driven by a reconstruction objective, requires large in-domain corpora ($\sim$184k sentences) but yields the most discriminative global score distributions, hence the best detection. CSE, driven by a contrastive objective, attains its optimum with only $4$k--$8$k raw in-domain sentences---making per-corpus or per-document adaptation realistic with no annotation\footnote{A full CSE training run on 4k--8k raw sentences completes in a few tens of seconds on a single laptop GPU (NVIDIA RTX PRO 4000, Blackwell generation).}---and produces spaces that excel at retrieval. This division of labour is what the two objectives predict (\autoref{sec:cse}): the contrastive loss acts directly on the cosine similarities used for ranking, whereas the reconstruction loss constrains them only indirectly, consistent with the alignment/uniformity account of contrastive learning \citep{wang2020alignment} and with the measurements of \autoref{sec:disc_umap}. The transfer results of \autoref{sec:results_latin_gen} further show that these adaptations generalize across works and authors within patristic Latin, and \autoref{sec:results_greek} that the CSE gains over the base models are of similar size on one Ancient Greek work. More specifically, however, we observed that in contrastive learning, the greater the deviation of the experimental data from the training data of language models, the greater the need for linguistic adaptation. While TSDAE shows a consistent ability to generalize over all types of data in detection, CSE adapted encoders obtain notably higher scores, especially in retrieval, when the training is performed directly on sentences taken from the experimental texts, even if they are noisier.           

\subsection{Geometry of the learned spaces}
\label{sec:disc_umap}

To relate the gains to the geometry of the representation spaces, we measure, for every Latin encoder of \autoref{tab:trin_results} except LatinBERT, the anisotropy, alignment, and uniformity of the joint embeddings of \textit{Vulgate} verses and \textit{De Trinitate} sentences, together with a standardized gold-pair margin (definitions and values in \autoref{app:geometry}). Contrastive adaptation makes the space markedly more isotropic than its base model (mean cosine similarity of random pairs 0.26 against 0.46) and more uniform ($\ell_{\mathrm{unif}}$ $-2.89$ against $-2.08$). It does not bring gold pairs closer in absolute terms ($\ell_{\mathrm{align}}$ 0.86 against 0.67), since it never sees them, but it spreads the rest of the space far more, so that the margin of gold pairs over random pairs doubles (6.0 against 3.1 standard deviations). TSDAE moves the other way: the space becomes more anisotropic (0.67) and less uniform ($-1.28$); gold pairs are closer in absolute terms (0.40) only because all pairs are, and the margin is essentially unchanged (3.2). Across the eight encoders, the margin orders the models almost as $Hits@1$ does (Spearman $\rho = 0.90$, computed on unrounded values), whereas anisotropy alone does not ($\rho = -0.62$): LaBSE and the fine-tuned multilingual-e5 have the same anisotropy (0.60) but lie at opposite ends of the retrieval ranking. Retrieval quality is thus a matter of relative rather than absolute proximity, and isotropy helps insofar as it widens the gap between related and unrelated pairs. These measurements bear out the prediction of \autoref{sec:cse}: the objective that acts directly on cosine similarities reshapes the space used for ranking, the reconstruction objective does not.

\begin{figure}[htbp]
    \centering
    \includegraphics[width=0.75\linewidth]{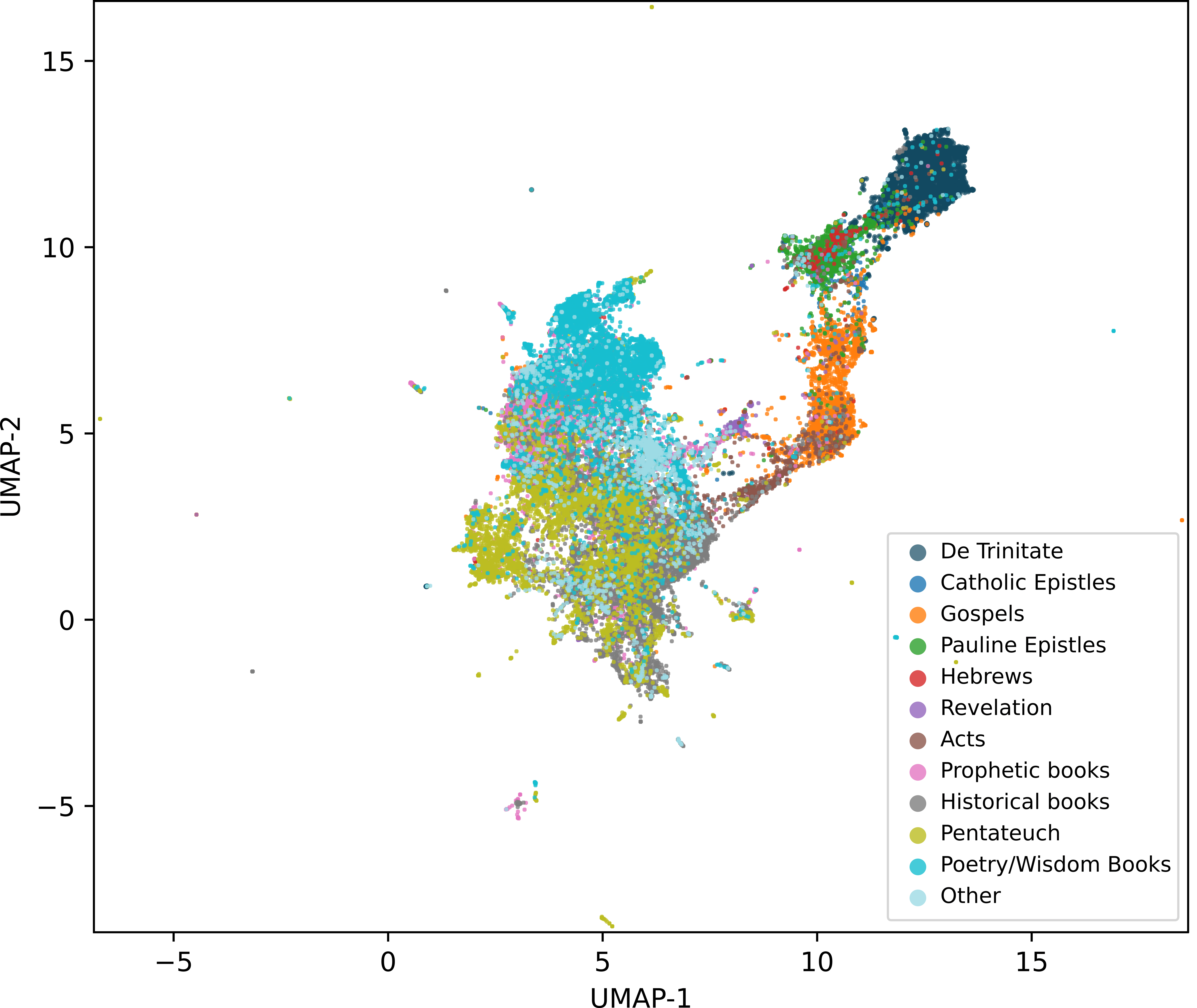}\\[0.5em]
    \includegraphics[width=0.75\linewidth]{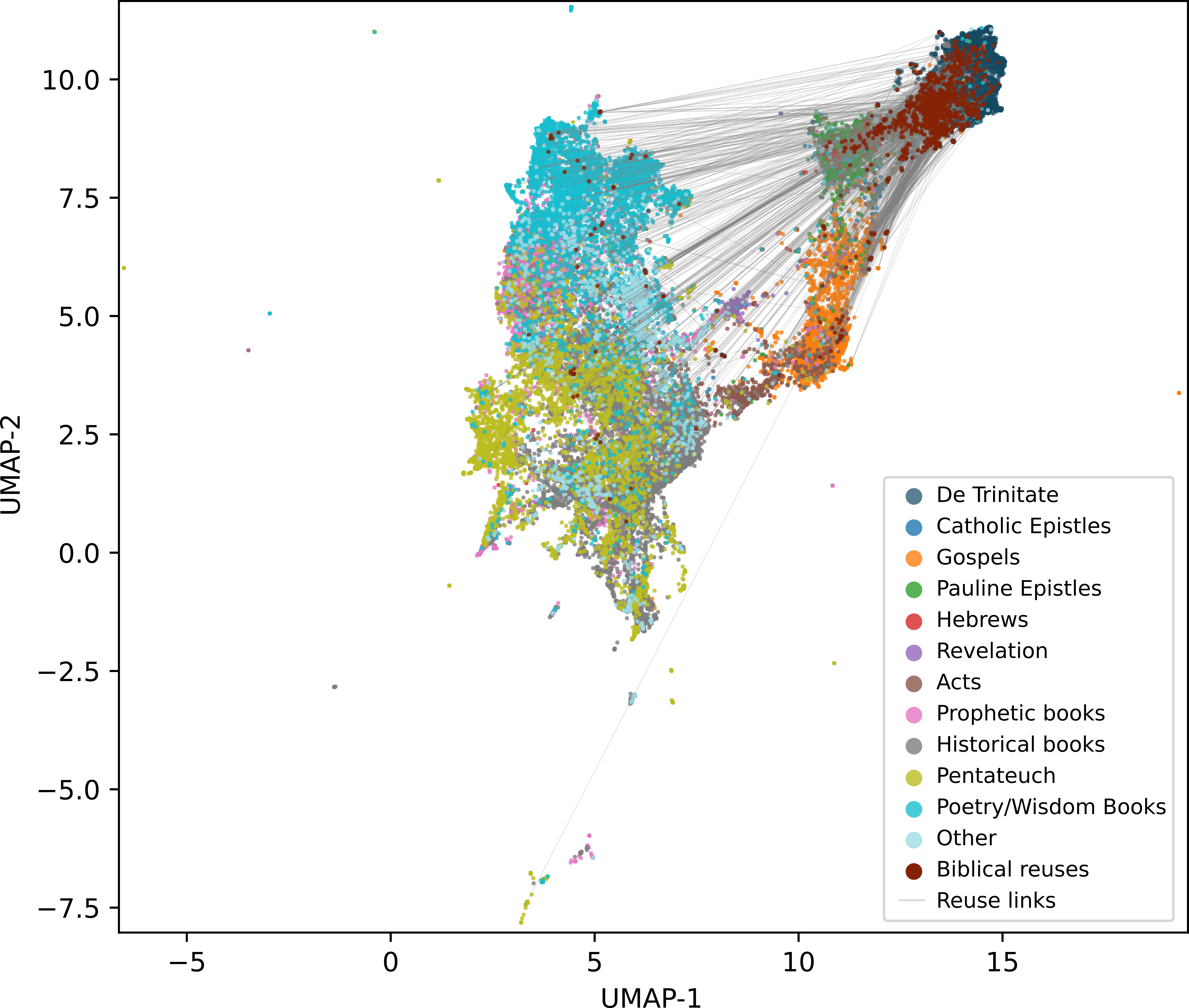}
    \caption{UMAP projection of biblical verses and \textit{De Trinitate} sentence representations from the LaBerta encoder. Each point is a sentence or a verse. Top: points colored by source. Bottom: same projection with gold reuse data superposed; \textit{De Trinitate} sentences containing biblical reuses are shown as dark red dots, with gray lines pointing to the linked verses.}
    \label{fig:umap_laberta}
\end{figure}

\begin{figure}[htbp]
    \centering
    \includegraphics[width=0.75\linewidth]{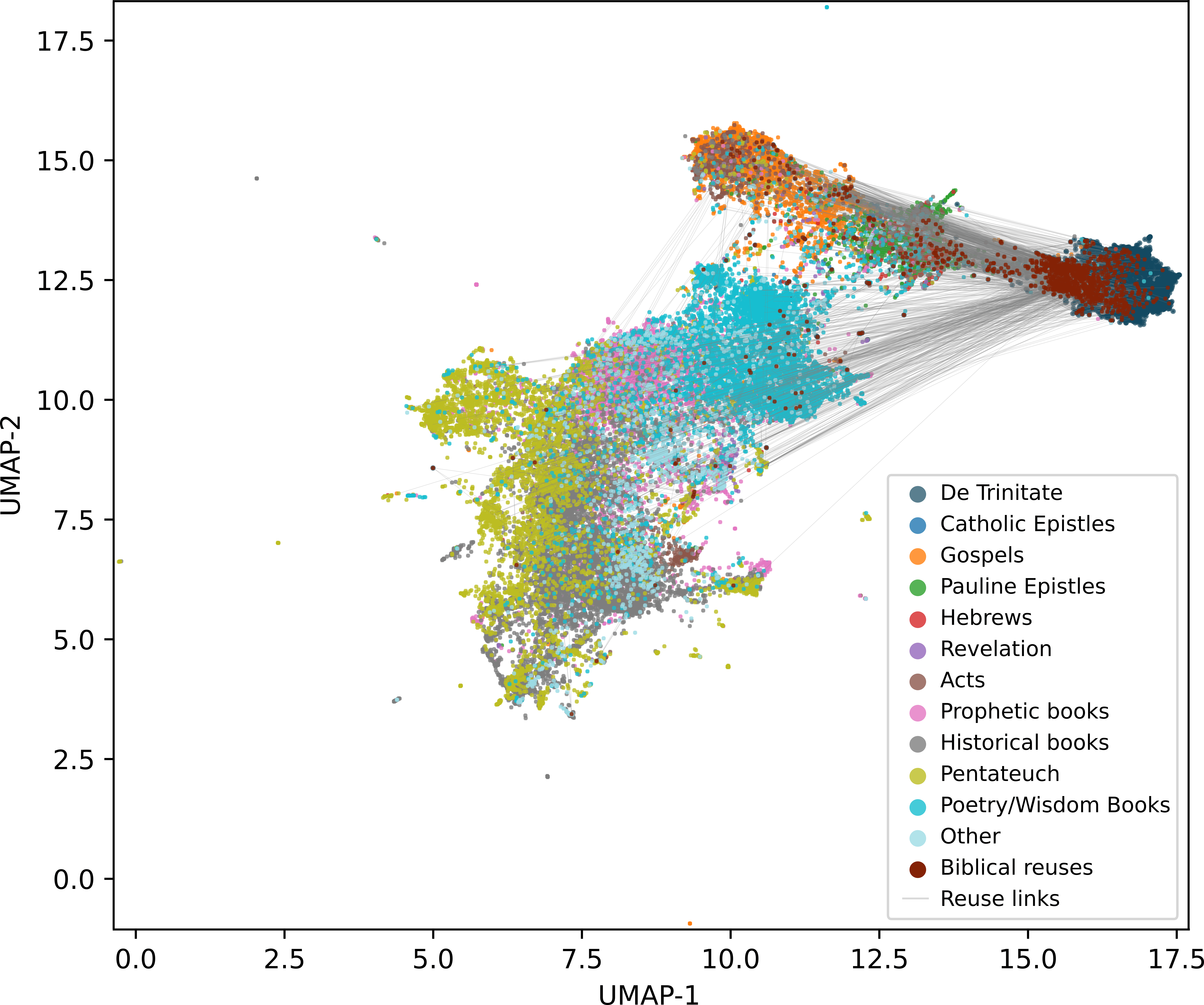}\\[0.5em]
    \includegraphics[width=0.75\linewidth]{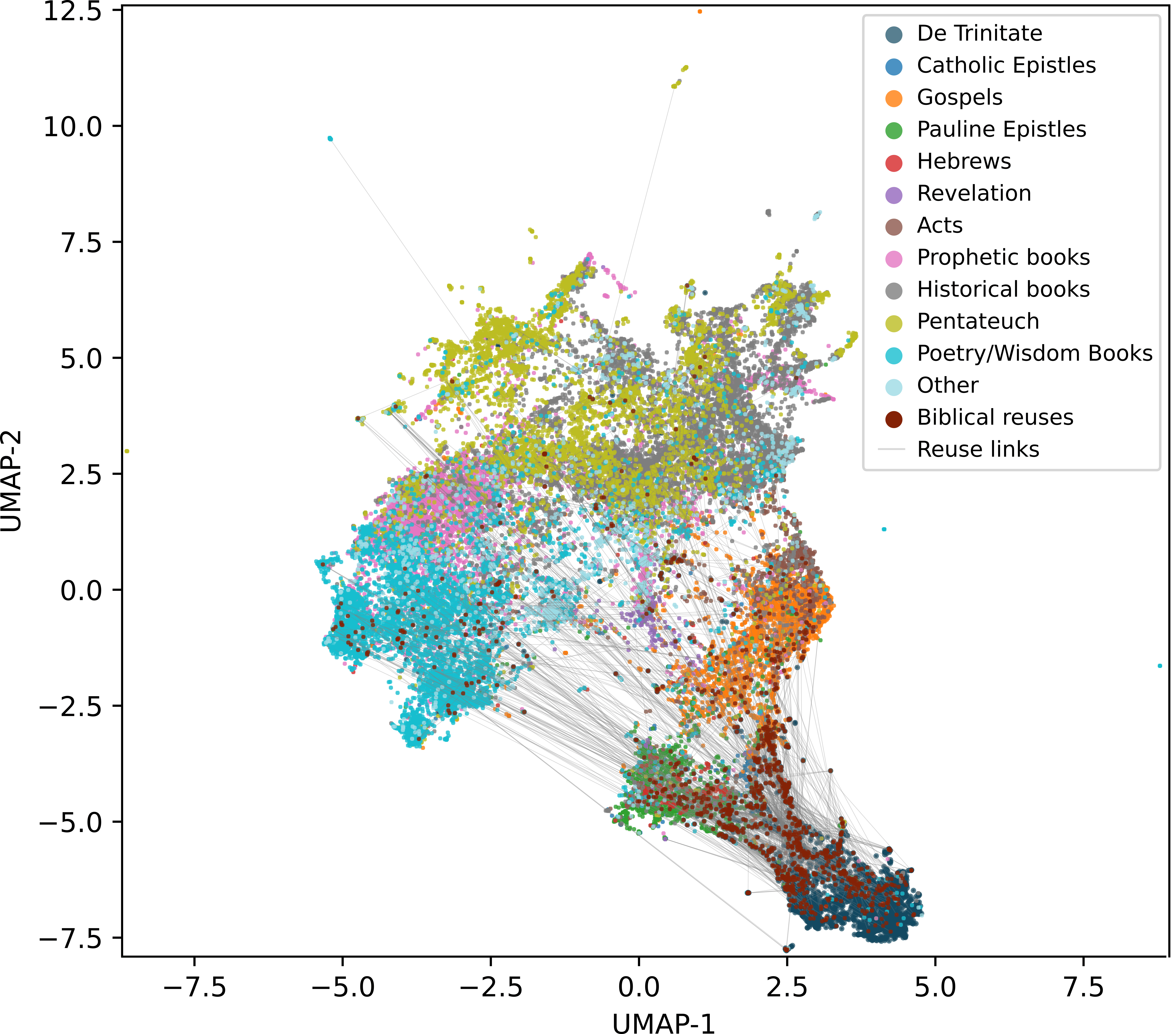}
\caption{UMAP projection of biblical verses and \textit{De Trinitate} sentence representations from the LaBerta-TSDAE encoder (top) and the LaBerta-CSE\textsubscript{trin} encoder (bottom). \textit{De Trinitate} sentences containing biblical reuses are shown as dark red dots, with gray lines pointing to the linked verses.}
    \label{fig:umap_tsdae-cse}
\end{figure}

\paragraph{UMAP atlases as exploratory views.} As a qualitative complement, we build 2D UMAP \citep{mcinnes2018umap} projections---so-called ``atlases''---of the joint embeddings of biblical verses and \textit{De Trinitate} sentences for the two adapted encoders and their base model LaBerta (Figures~\ref{fig:umap_laberta}--\ref{fig:umap_tsdae-cse}). Each sentence or verse is represented by a dot colored by its source; in the panels with gold data, \textit{De Trinitate} sentences carrying a reuse are shown as dark red dots, with gray lines pointing to their verses. UMAP preserves local neighbourhoods but not global distances: the distances between clusters and the lengths of links drawn across the projection could not be faithful to the embedding space \citep{mcinnes2018umap,chari2023specious}. We therefore do not read link lengths or cluster separations as evidence. Of the impressions the atlases suggest, that CSE brings reuse sentences closer to their verses holds in relative terms only, as the gold-pair margin shows (\autoref{tab:geometry}). The atlases show a local organization that is philologically coherent: points group by source, biblical books group in a way consistent with the bipartition between Old and New Testament, and \textit{De Trinitate} sentences lie near the Pauline epistles and Hebrews. Their main use is exploratory: a scholar can navigate them interactively to find thematic regions and clusters of related verses and sentences.

\section{Limitations}
\label{sec:limitations}

Several limitations qualify our conclusions.
\begin{enumerate}
    \item \textbf{Tested envelope.} The gold data is restricted to biblical reuse in quotation-dense patristic prose, matched against a closed canonical source corpus, with positive rates of 23\% to 50\%. Whether the adaptation gains hold for allusive poetry, where a connection may rest on a single rare word, a sound pattern, or a syntactic frame, for classical intertextuality more generally, for documentary or fragmentary corpora, and for other ancient languages remains to be tested. Ancient Greek is represented by a single work and 106 reuse sentences.
    \item \textbf{Simulated noise.} The robustness experiments perturb a clean edition with synthetic character- and token-level noise (\autoref{sec:noise}). They do not reproduce line-segmentation, reading-order, and layout errors, nor the expansion policies of real transcription pipelines, so the noise results are not tested properties of ATR output.
    \item \textbf{Sentence-level segmentation.} Since the text is segmented at the sentence level, some sentences contain several reuses. Because the detection label states ``at least one reuse in this sentence'', a sentence containing both an allusion and a quotation counts as detected if either is found; this applies, for example, to 44 of the 269 allusions in the \textit{Adversus Iovinianum}.
    \item \textbf{Hyper-parameters.} The sensitivity of both strategies to their hyper-parameters has not been explored exhaustively, especially learning rates, weight decay, the SimCSE temperature, and the TSDAE token deletion ratio, for which default values were used.
\end{enumerate}

\section{Conclusion and Perspectives}
\label{sec:conclusion}

This article addressed the analytical step that follows automatic transcription: enabling semantic, cross-corpus analysis of ancient-language texts through sentence representations adapted to the corpus under study. Working on the philologically central case of biblical reuse in patristic literature---2{,}936 sentences carrying expert-verified reuses in Latin and Ancient Greek---it makes five contributions. (i)~\textbf{Methodologically}, we decompose text reuse identification into two separately evaluated tasks, detection (binary classification) and correspondence retrieval (IR), and show that they should not be conflated: on every Latin test set the detection leader and the retrieval leader differ, and retrieval is the more discriminating evaluation. (ii)~We \textbf{benchmark}, on both tasks, multilingual sentence encoders, specialized language models used as pooled sequence encoders, and distilled or supervised fine-tuned sentence encoders, exposing the limitations of each family for ancient languages. (iii)~We show that two \textbf{fully unsupervised strategies}, TSDAE and CSE, adapt a language model into a corpus-specific sequence encoder. Taken together, they bring substantial gains on both tasks, with complementary profiles---TSDAE for detection only, given a large in-domain corpus ($\sim$184k sentences)\footnote{This Latin sentence embedding model is available at \url{https://huggingface.co/TdelaSelle/PatriLaSE}}; CSE for retrieval, where it leads by a wide margin, and to a lesser extent for detection, at an extremely light cost ($\approx$4--8k raw sentences and a few tens of seconds on a laptop GPU)---hence applicable corpus by corpus, without any annotation, including, under simulated noise, as a remediation step trained directly on the perturbed text. (iv)~\textbf{Measurements of alignment, uniformity, and anisotropy} connect the geometric effect of each training strategy to the measured gains, while UMAP atlases serve as exploratory instruments for philological work. (v)~These methods are made available to non-specialists through \textit{Paraphrasis} \citep{swh-dir-b0dd647}, an online tool covering raw-text segmentation, model fine-tuning, and cross-corpus semantic search.

Several perspectives follow naturally. The most immediate is to test the method outside the present envelope: on allusive poetry and classical intertextuality, for which public benchmarks now exist \citep{dexter2024database,schelb_loci_2026}, on further works, authors, and languages, on real post-ATR output, and on other applications of sentence embeddings such as text alignment and authorship attribution. A second direction is methodological: designing and evaluating a two-stage pipeline that couples a detector and a retriever, possibly built from different encoders; measuring, within the same domain, how many labeled pairs a supervised encoder needs to match unsupervised adaptation, exploring the hyper-parameter sensitivity left open here, mitigating sentence-level segmentation biases, and assembling larger premodern Greek corpora so that reconstruction-based adaptation becomes feasible beyond Latin. Finally, text reuse identification methods would benefit from larger and more diverse biblical reuse datasets for supervised learning and refined model evaluation, leveraging the digitization of patristic works and scriptural apparatus.

\paragraph{Data and code availability.} The code, data, and models supporting the results are released as follows.
\begin{itemize}
    \item \textbf{Code and configurations}: text preprocessing, training, evaluation, noise-simulation, and analysis scripts, in the public repository \url{https://github.com/TdelaSelle/Patristic_sentence_representation}, under the MIT licence. The release used for this article is archived on Zenodo, \url{https://doi.org/10.5281/zenodo.24166226}.
    \item \textbf{Gold annotations and Biblical or Patristic texts}: The patristic texts were obtained as files from the CAG (Augustine) and from BibliText (Jerome and Athanasius; \autoref{tab:corpora}); most of these texts are the property of Brepols Publishers and were used under licence, so they cannot be shared publicly. Equivalent access may be obtained through a Brepols subscription. The biblical reference data used in this study come from the BiblIndex platform (\url{https://www.biblindex.org/en/bible-parser}); although accessible via the BiblIndex interface and API, they are not freely redistributable under the platform's terms of use. Only the open-access data derived from Augustine of Hippo's \textit{Contra Faustum} (public-domain text) are deposited in our repository.
    \item \textbf{Models}: the Latin TSDAE model PatriLaSE is available at \url{https://huggingface.co/TdelaSelle/PatriLaSE} under the CC-BY 4.0 licence.
    \item \textbf{Tool}: the \textit{Paraphrasis} \citep{swh-dir-b0dd647} tool is permanently archived in the Software Heritage archive under the persistent identifier SWHID \nolinkurl{swh:1:snp:94bb2eb096d6820dbaf1fe34dcd83b1eee4d5aa2} and is accessible at \url{https://tdelaselle.gitpages.huma-num.fr/paraphrasis/} (MIT license).
\end{itemize}

\paragraph{Acknowledgements.} This research received funding from Biblissima+, Observatoire des cultures écrites anciennes, de l'argile à l'imprimé - ANR-21-ESRE-0005, \url{https://projet.biblissima.fr/en}. A CC-BY public copyright license has been applied by the authors to the present document and will be applied to all subsequent versions up to the Author Accepted Manuscript arising from this submission, in accordance with the grant’s open access conditions.

\appendix

\section{Simulated noise: HTR and abbreviations}\label{app:noise}

Both noise models below are forward simulations applied to a clean edition. They are calibrated on published error rates and on scribal conventions, not fitted to the recognition output of a particular model on the manuscripts of these works, and they leave out segmentation and layout errors (\autoref{sec:noise}).

\textbf{HTR noise.} To simulate the output of Handwritten Text Recognition systems applied to medieval Latin manuscripts, we use a forward (clean → noisy) character-level noise model calibrated to a target Character Error Rate (CER), set to 5\% in our experiments \citep{clerice2024catmus}.
Errors are drawn from an empirically motivated taxonomy: single-character substitutions from a paleographic confusion matrix dominate (roughly 70\% of errors), chiefly minim confusions typical of Latin minuscule (\textit{n}$\leftrightarrow$\textit{u}, \textit{m}→\textit{in}/\textit{ni}, \textit{i}$\leftrightarrow$\textit{l}), followed by multi-character substitutions (15\%, e.g., \textit{in}→\textit{m}, \textit{cl}→\textit{d}), character deletions (7\%), insertions of minim-class letters (4\%), word-boundary errors (3\%, spaces dropped or added), and residual unexpanded abbreviations (1\%, e.g., \textit{deus}→\textit{ds}, \textit{-que}→\textit{q;}). Each sentence is corrupted with a deterministic per-sentence seed for reproducibility and the achieved CER is verified post hoc via Levenshtein distance.

\textbf{Abbreviation noise.} A second forward noise model simulates medieval scribal abbreviation conventions. Abbreviations are applied stochastically per occurrence — so that full and abbreviated forms coexist within a text, as in real manuscripts — in five ordered classes with class-specific probabilities: multi-word syntagms (\textit{spiritus sanctus}→\textit{sps scs}, p=0.75), nomina sacra (\textit{deus}→\textit{ds}, \textit{christus}→\textit{xps}, p=0.85, reflecting their near-universal abbreviation), standard contractions of high-frequency words (\textit{autem}→\textit{aut}, \textit{quod}→\textit{qd}, p=0.55), \textit{per-}/\textit{pro-}/\textit{prae-} prefix signs (p=0.30–0.40), and final-letter suspensions with conventional plain-text renderings of the manuscript signs (nasal \textit{-m}→\textit{\~}, \textit{-us}→\textit{9}, \textit{-orum}→\textit{oz}, \textit{-bus}→\textit{b;}, \textit{-que}→\textit{q;}, p=0.20–0.70), plus the Tironian \textit{et}→\textit{7} (p=0.30). The per-class rates are calibrated so that roughly 25–35\% of tokens are abbreviated overall. As with the HTR model, corruption is seeded per sentence and the resulting character reduction is measured empirically.

\section{Corpus-size experiments}
\label{app:corpus_size}

This appendix reports how the two adaptation strategies respond to the size of the (unlabeled) training corpus, in Latin and in Greek. For Latin (\autoref{tab:corpus_sizes}) we vary the number of raw sentences by random sampling---from the patristic corpus for TSDAE, from the Augustinian corpus for CSE---and evaluate on the \textit{De Trinitate}. For Greek (\autoref{tab:greek_corpus_sizes}) we vary the contrastive training size for Logion-CSE and evaluate on the \textit{De decretis Nicaenae synodi}.

\begin{table}[htbp]
    \centering
    \caption{Effect of training-corpus size on the \textit{De Trinitate} (\textsc{Augustine of Hippo}): text reuse detection ($AP$, $AUC$-$ROC$, $F1_{\max}$) and correspondence retrieval against the \textit{Vulgate} ($Hits@1$, $Hits@10$; over $N = 1189$ reuse sentences). Subscripts give the training-corpus size in sentences.}
    \label{tab:corpus_sizes}
    \begin{tabular}{lrrr|rr}
    \toprule
    Model & $AP$ & $AUC$-$ROC$ & $F1_{\max}$ & $Hits@1$ & $Hits@10$\\
    \midrule
    LaBerta-TSDAE\textsubscript{184k} & \textbf{0.812} & \textbf{0.900} & \textbf{0.731} & 0.401 & 0.598\\
    LaBerta-TSDAE\textsubscript{40k}  & 0.751 & 0.861 & 0.670 & 0.374 & 0.542\\
    LaBerta-TSDAE\textsubscript{5k}   & 0.429 & 0.618 & 0.445 & 0.213 & 0.309\\
    LaBerta-CSE\textsubscript{Aug}  & 0.767 & 0.848 & 0.677 & 0.596 & 0.752\\
    LaBerta-CSE\textsubscript{9k}   & 0.797 & 0.868 & 0.718 & 0.603 & 0.759\\
    LaBerta-CSE\textsubscript{8k}   & 0.802 & 0.874 & 0.726 & 0.609 & 0.764\\
    LaBerta-CSE\textsubscript{6k}   & 0.800 & 0.872 & 0.722 & \textbf{0.611} & 0.765\\
    LaBerta-CSE\textsubscript{4k}   & 0.799 & 0.871 & 0.722 & 0.609 & \textbf{0.770}\\
    LaBerta                         & 0.717 & 0.822 & 0.624 & 0.424 & 0.560\\
    \bottomrule
    \end{tabular}
\end{table}

\begin{table}[htbp]
    \centering
    \caption{Effect of contrastive training-corpus size for Logion-CSE on the \textit{De decretis Nicaenae synodi} (\textsc{Athanasius of Alexandria}): detection ($AP$, $AUC$-$ROC$, $F1_{\max}$) and retrieval ($Hits@1$, $Hits@10$; over $N = 106$ reuse sentences). Logion is the unadapted base model.}
    \label{tab:greek_corpus_sizes}
    \begin{tabular}{lrrr|rr}
    \toprule
    Model & $AP$ & $AUC$-$ROC$ & $F1_{\max}$ & $Hits@1$ & $Hits@10$\\
    \midrule
    Logion-CSE\textsubscript{9k} & 0.734 & 0.834 & 0.670 & 0.387 & \textbf{0.575}\\
    Logion-CSE\textsubscript{8k} & \textbf{0.737} & 0.837 & \textbf{0.681} & 0.387 & \textbf{0.575}\\
    Logion-CSE\textsubscript{6k} & 0.736 & \textbf{0.842} & 0.678 & \textbf{0.396} & \textbf{0.575}\\
    Logion-CSE\textsubscript{5k} & 0.735 & \textbf{0.842} & \textbf{0.681} & 0.387 & 0.566\\
    Logion                       & 0.687 & 0.812 & 0.603 & 0.283 & 0.377\\
    \bottomrule
    \end{tabular}
\end{table}

The two strategies exhibit opposite data regimes. \textbf{Contrastive adaptation (CSE)} reaches optimal performance with \emph{very small, highly specialized} corpora of $[4\mathrm{k}, 8\mathrm{k}]$ sentences---a training set attainable at the scale of a single work. Performance is essentially flat across $4$k--$9$k (retrieval $Hits@1$ $0.603$--$0.611$) and, notably, enlarging the corpus to the full Augustinian collection (CSE\textsubscript{Aug}) slightly \emph{degrades} retrieval ($Hits@1$ $0.596$ vs.\ $0.611$ at $6$k), indicating that domain specificity matters more than sheer data quantity for contrastive adaptation. Owing to the very small training sets involved, contrastive adaptation is also computationally trivial: a full CSE training run on 4k--8k raw sentences completes in a few tens of seconds on a single laptop GPU (NVIDIA RTX PRO 4000, Blackwell generation). \textbf{TSDAE} achieves its best detection only with a large specialized corpus (the full patristic corpus, $\sim$184k deduplicated sentences) and collapses as corpus size decreases (AUC-ROC 0.618 at 5k, below the unadapted LaBerta), reflecting the data appetite of a reconstruction objective that must learn to structure the space from scratch. The Greek experiment (\autoref{tab:greek_corpus_sizes}) reproduces the CSE saturation: Logion-CSE is nearly invariant across 5k--9k sentences (all within $0.008$ $AUC$-$ROC$ and, given $N = 106$, within sampling noise on retrieval), while every size improves markedly over the unadapted Logion. This contrast is practically important: CSE offers a near-instant, per-corpus adaptation, whereas TSDAE is better reserved for settings where a large in-domain corpus is available.

\section{Comparison with a non-neural baseline}
\label{app:results_other}

As a non-neural baseline, we run \href{https://github.com/dasmiq/passim}{Passim} — the reference n-gram-indexing, Smith-Waterman-alignment text-reuse detector — over the same \textit{De Trinitate} and \textit{Vulgate} data. With its \textbf{default parameters} (\verb|n|=25-word shingles, \verb|m|=1 minimum matching n-gram, 50-character minimum alignment), Passim is highly precise but severely under-recalls at verse granularity (see the first line of \autoref{tab:other_systems}), since 25-word shingles rarely survive intact across paraphrastic biblical reuse. We therefore \textbf{grid-search} the two canonical levers — n-gram order \verb|n| $\in$ \{5, 8, 10, 12, 15, 20, 25\} and minimum matching n-grams \verb|m| $\in$ \{1, 2\} — with the minimum alignment length lowered to 20 characters to accommodate short verses, selecting the configuration that maximizes $Hits@1$. The optimum sharply improves F1 and $Hits@1$ scores — still well below the embedding-based models, which is expected since Passim's shingle-matching approach cannot capture paraphrastic or semantic reuse (allusions, inexact quotations)  \citep{manjavacas2019allusive} the way dense sentence embeddings can. 
\begin{table}[htbp]
    \centering
    \caption{Biblical reuse identification on the \textit{De Trinitate} (\textsc{Augustine of Hippo}): detection (\textit{Precision}, \textit{Recall}, F1) and retrieval ($Hits@1$) with Passim. For the optimized version (\textbf{n=5, m=1-2, min-align=20}, m=1 and m=2 gave identical F1), Passim's parameters have been tuned through grid search.}
    \label{tab:other_systems}
    \begin{tabular}{l|rrr|r}
    \toprule
    System & \textit{Precision} & \textit{Recall} & F1 & $Hits@1$\\
    \midrule
    Passim (default parameters)& 0.99 & 0.06 & 0.11 & 0.06 \\
    Passim (grid search optimization)& 0.50 & 0.44 & 0.47 & 0.25\\
    \bottomrule
    \end{tabular}
\end{table}

\section{Variability: bootstrap intervals and training seeds}
\label{app:variability}

Two sources of variability affect the scores of \autoref{sec:results}: the sampling of test sentences, and the stochasticity of training. Both are estimated on the \textit{De Trinitate}.

\paragraph{Sampling of test sentences.} \autoref{tab:bootstrap} gives, for every model of \autoref{tab:trin_results} except LatinBERT, a 95\% percentile bootstrap interval for each metric (1{,}000 resamples of the 4{,}410 sentences for detection, of the 1{,}189 reuse sentences for retrieval) \citep{efron1993bootstrap}. Half-widths are $\pm0.01$ to $\pm0.03$ for detection and about $\pm0.03$ for $Hits@k$. The retrieval lead of LaBerta-CSE\textsubscript{trin} lies far outside them: its $Hits@1$ interval, [0.58, 0.63], is disjoint from that of every other model (at most [0.46, 0.51]). Both adapted encoders are separated from LaBerta on every detection metric. LaBerta-TSDAE and LaBerta-CSE\textsubscript{trin} are separated on $AUC$-$ROC$ but not on $AP$ or $F1_{\max}$. Disjoint intervals indicate a difference beyond sampling error; overlapping intervals do not by themselves establish that there is none.

\begin{table}[htbp]
    \centering
    \caption{95\% percentile bootstrap intervals over test sentences on the \textit{De Trinitate} (1{,}000 resamples; single training run). Value [lower bound, upper bound].}
    \label{tab:bootstrap}
    \resizebox{\textwidth}{!}{%
    \begin{tabular}{lccc|cc}
    \toprule
    Model & $AP$ & $AUC$-$ROC$ & $F1_{\max}$ & $Hits@1$ & $Hits@10$\\
    \midrule
    LaBerta-TSDAE & 0.81 [0.79, 0.83] & 0.90 [0.89, 0.91] & 0.73 [0.71, 0.75] & 0.40 [0.37, 0.43] & 0.60 [0.57, 0.63]\\
    LaBerta-CSE\textsubscript{trin} & 0.79 [0.77, 0.82] & 0.87 [0.85, 0.88] & 0.71 [0.69, 0.74] & 0.61 [0.58, 0.63] & 0.75 [0.73, 0.78]\\
    LaBerta & 0.72 [0.69, 0.74] & 0.82 [0.81, 0.84] & 0.62 [0.61, 0.65] & 0.42 [0.39, 0.45] & 0.56 [0.53, 0.59]\\
    SPhilBerta-emb-lat-intertext-v1 & 0.66 [0.64, 0.69] & 0.79 [0.78, 0.81] & 0.58 [0.56, 0.60] & 0.43 [0.40, 0.46] & 0.59 [0.56, 0.61]\\
    SPhilBerta & 0.62 [0.60, 0.65] & 0.77 [0.76, 0.79] & 0.55 [0.53, 0.58] & 0.39 [0.36, 0.42] & 0.56 [0.53, 0.58]\\
    multilingual-e5-base-emb-lat-intertext-v1 & 0.65 [0.63, 0.68] & 0.76 [0.75, 0.78] & 0.56 [0.54, 0.59] & 0.48 [0.46, 0.51] & 0.64 [0.61, 0.67]\\
    multilingual-e5-base & 0.60 [0.57, 0.63] & 0.73 [0.71, 0.75] & 0.52 [0.50, 0.55] & 0.37 [0.34, 0.39] & 0.50 [0.48, 0.53]\\
    LaBSE & 0.52 [0.50, 0.55] & 0.70 [0.68, 0.71] & 0.50 [0.48, 0.52] & 0.30 [0.27, 0.32] & 0.41 [0.39, 0.44]\\
    \bottomrule
    \end{tabular}}
\end{table}

\paragraph{Training seeds.} LaBerta-CSE\textsubscript{trin} was retrained with three random seeds (\autoref{tab:seeds}). Standard deviations range from 0.013 ($AUC$-$ROC$) to 0.029 ($Hits@1$), the same order as the bootstrap half-widths: training stochasticity adds a variability comparable to that of test sampling. Measured against these standard deviations, the retrieval lead of LaBerta-CSE\textsubscript{trin} over the next model ($0.13$ in $Hits@1$, \autoref{tab:trin_results}) is large, whereas its detection gap to LaBerta-TSDAE exceeds seed variability on AUC-ROC only (0.03 against a standard deviation of 0.013), and is of the order of the standard deviations on AP and F1max.

\begin{table}[htbp]
    \centering
    \caption{Training-seed variability of LaBerta-CSE\textsubscript{trin} on the \textit{De Trinitate}: standard deviation of each metric over three training seeds.}
    \label{tab:seeds}
    \small
    \begin{tabular}{lccc|cc}
    \toprule
    Model & $AP$ & $AUC$-$ROC$ & $F1_{\max}$ & $Hits@1$ & $Hits@10$\\
    \midrule
    LaBerta-CSE\textsubscript{trin} (SD) & 0.018 & 0.013 & 0.026 & 0.029 & 0.017\\
    \bottomrule
    \end{tabular}
\end{table}

\section{Geometry of the representation spaces}
\label{app:geometry}

The measures are computed on the joint set of \textit{Vulgate} verses and \textit{De Trinitate} sentences, all embeddings being L2-normalized. \emph{Anisotropy} is the mean cosine similarity between random pairs of embeddings \citep{ethayarajh2019contextual}. \emph{Alignment}, $\ell_{\mathrm{align}} = \mathbb{E}_{(x,y)} \lVert f(x) - f(y) \rVert_2^2$, is computed over the 1{,}899 gold (sentence, verse) pairs, and \emph{uniformity}, $\ell_{\mathrm{unif}} = \log \mathbb{E}_{x,y} \exp\bigl(-2 \lVert f(x) - f(y) \rVert_2^2\bigr)$, over random pairs \citep{wang2020alignment}. Lower anisotropy and $\ell_{\mathrm{unif}}$ indicate a more isotropic and more uniform space; lower $\ell_{\mathrm{align}}$ indicates closer gold pairs, but a global contraction of the space lowers it as well, so it must be read together with uniformity. The \emph{gold-pair margin} summarizes the two: the mean cosine similarity of gold pairs minus that of random sentence--verse pairs, divided by the standard deviation of the latter; unlike $\ell_{\mathrm{align}}$, it is insensitive to a global contraction or dilation of the space.

\autoref{tab:geometry} reports the measures for the eight encoders, with their $Hits@1$ from the same run. Besides the contrast between CSE and TSDAE discussed in \autoref{sec:disc_umap}, two observations stand out. The most anisotropic encoder, multilingual-e5-base (0.83), has the lowest $\ell_{\mathrm{align}}$ (0.25) only because all its pairs are close, which illustrates why alignment cannot be read alone. Supervised fine-tuning on Latin parallels also widens the margin (multilingual-e5-base $2.6 \rightarrow 3.8$; SPhilBerta $2.6 \rightarrow 3.1$), but far less than unsupervised contrastive adaptation of LaBerta ($3.1 \rightarrow 6.0$). Over the eight encoders, the rank correlation with $Hits@1$ is $\rho = 0.90$ for the margin and $\rho = -0.62$ for anisotropy; eight models are too few for these coefficients to carry more than a descriptive value.

\begin{table}[htbp]
    \centering
    \caption{Geometry of the representation spaces (\textit{Vulgate} verses and \textit{De Trinitate} sentences, L2-normalized embeddings). Anisotropy: mean cosine of random pairs; $\ell_{\mathrm{align}}$ over gold pairs and $\ell_{\mathrm{unif}}$ over random pairs \citep{wang2020alignment}; margin: standardized difference between the mean cosine of gold pairs and that of random sentence--verse pairs. $Hits@1$ from the same run, for reference.}
    \label{tab:geometry}
    \small
    \begin{tabular}{lrrrr|r}
    \toprule
    Model & Anisotropy $\downarrow$ & $\ell_{\mathrm{align}}$ & $\ell_{\mathrm{unif}}$ $\downarrow$ & Margin $\uparrow$ & $Hits@1$\\
    \midrule
    LaBerta-TSDAE & 0.67 & 0.40 & $-$1.28 & 3.2 & 0.40\\
    LaBerta-CSE\textsubscript{trin} & 0.26 & 0.86 & $-$2.89 & 6.0 & 0.61\\
    LaBerta & 0.46 & 0.67 & $-$2.08 & 3.1 & 0.42\\
    SPhilBerta-emb-lat-intertext-v1 & 0.53 & 0.54 & $-$1.82 & 3.1 & 0.43\\
    SPhilBerta & 0.51 & 0.57 & $-$1.88 & 2.6 & 0.39\\
    multilingual-e5-base-emb-lat-intertext-v1 & 0.60 & 0.58 & $-$1.61 & 3.8 & 0.48\\
    multilingual-e5-base & 0.83 & 0.25 & $-$0.67 & 2.6 & 0.37\\
    LaBSE & 0.60 & 0.58 & $-$1.55 & 1.7 & 0.30\\
    \bottomrule
    \end{tabular}
\end{table}

\printbibliography

\end{document}